\documentclass[review]{elsarticle}

\usepackage{graphicx} 
\graphicspath{{fig/}}
\usepackage{adjustbox} 
\usepackage{color} 
\usepackage{enumerate} 
\usepackage{geometry} 
\usepackage{amsmath} 
\usepackage{amssymb} 
\usepackage{eurosym} 
\usepackage[mathletters]{ucs} 
\usepackage[OT4]{fontenc}
\usepackage[utf8]{inputenc}

\usepackage{float}
\usepackage{fancyvrb} 
\usepackage{grffile}
\usepackage{hyperref}
\usepackage{longtable} 
\usepackage{booktabs}  
\usepackage{multirow}
\usepackage{numprint}
\npthousandsep{\,}

\usepackage[figure,vlined,linesnumbered]{algorithm2e}

\usepackage{tikz}
\usepackage{layouts}

\usepackage{subcaption}

\definecolor{darkorange}{rgb}{.71,0.21,0.01}
\definecolor{darkgreen}{rgb}{.12,.54,.11}
\definecolor{darkred}{rgb}{.54,.1,.1}
\definecolor{gray}{gray}{0.45}
\definecolor{blue}{rgb}{0,.145,.698}

\newcommand{\R}[1]{{\color{black} #1}}

\hypersetup{
  breaklinks=true,  
  colorlinks=true,
  urlcolor=blue,
  linkcolor=darkorange,
  citecolor=darkgreen,
}

\newcommand{\doi}[1]{\textsc{doi}: \href{http://dx.doi.org/#1}{\nolinkurl{#1}}}

\makeatletter
\newcommand\thefontsize{The current font size is: \f@size pt}
\makeatother

\makeatletter
\def\ps@pprintTitle{%
 \let\@oddhead\@empty
 \let\@evenhead\@empty
 \def\@oddfoot{Accepted Manuscript. DOI: 10.1016/j.asoc.2022.108653\hfill}%
 \let\@evenfoot\@oddfoot}
\makeatother

\begin{document}

\title{Improving Ant Colony Optimization Efficiency for Solving Large TSP Instances}
\author[uos]{Rafał Skinderowicz}
\ead{rafal.skinderowicz@us.edu.pl}

\address[uos]{University of Silesia, Institute of Computer Science,\\Będzińska
39, 41-205 Sosnowiec, Poland\\
\vspace{1em}
       {\rm
       \textcopyright 2022. This manuscript version is made available under the CC-BY-NC-ND 4.0 license http://creativecommons.org/licenses/by-nc-nd/4.0/
       }
}

\begin{abstract}
Ant Colony Optimization (ACO) is a family of nature-inspired metaheuristics
often applied to finding approximate solutions to difficult optimization
problems. Despite being significantly faster than exact methods, the ACOs
can still be prohibitively slow, especially if compared to basic
problem-specific heuristics. As recent research has shown, it is possible
to significantly improve the performance through algorithm refinements
and careful parallel implementation benefiting from multi-core CPUs and
dedicated accelerators. In this paper, we present a novel ACO variant,
namely the Focused ACO (FACO). One of the core elements of the FACO is a
mechanism for controlling the number of differences between a newly
    constructed and a selected previous solution. The mechanism results in a
more focused search process, allowing to find improvements while preserving
the quality of the existing solution. An additional benefit is a more
efficient integration with a problem-specific local search. 
Computational study based on a range of the Traveling Salesman
Problem instances shows that the FACO outperforms the state-of-the-art
ACOs when solving large TSP instances. Specifically, the FACO required less
than an hour of an 8-core commodity CPU time to find high-quality solutions
(within 1\% from the best-known results) for TSP Art Instances ranging from
    \numprint{100000} to \numprint{200000} nodes.
\end{abstract}


\maketitle
\section{Introduction}
\label{sec:Introduction}

Ant Colony Optimization (ACO) belongs to a growing collection of
nature-inspired metaheuristics that can be applied to solve various
optimization problems \cite{Dokeroglu2019, Yang2016}.  Heuristics, in general,
do not guarantee to find an optimum but can be helpful if the available
computational budget is insufficient to use an exact algorithm. Typically, this
is the case if one considers solving NP-hard combinatorial optimization
problems (COPs).  For example, even though the Traveling Salesman Problem (TSP) is one of
the most thoroughly studied COPs, still, to date, the largest solved to
optimality instance of the problem has only \numprint{85900} cities and the computations
took years of CPU time~\cite{Applegate2011}.

In general, the ACO refers not to a single algorithm but rather to a
\emph{family} of algorithms whose main principles mimic the behavior of certain species of
ants~\cite{Dorigo2004}.  Among the most well-known ACOs are the Ant
System~\cite{Colorni1991}, Ant Colony Sytem~\cite{Dorigo1997}, MAX-MIN Ant
System (MMAS)~\cite{Stutzle2000}, Population-based ACO~\cite{Guntsch2002}, and
Beam ACO~\cite{LopezIbanez2010}.  Since the first application of the Ant System
to solve the TSP~\cite{Colorni1991}, researchers have proposed numerous
applications of the ACOs including, among others, vehicle
routing~\cite{Bell2004}, the set cover problem~\cite{Leguizamon1999}, edge
detection on digital images~\cite{Nezamabadi2006}, protein
folding~\cite{Shmygelska2005}, and scheduling problems~\cite{Deng2019}.
The main motivations behind developing multiple ACO variants include adaptation
to new types of problems, improvements to the convergence to high-quality
solutions, and more efficient utilization of computing resources.
Additionally, some work was motivated by the ever-growing computational power
offered by multi-core CPUs~\cite{Pedemonte2011,Zhou2018}, specialized
accelerators~\cite{Martinez2021,Tirado2017}, and general-purpose graphics
processing units
(GPUs)~\cite{Cecilia2012,Delevacq2013,Skinderowicz2016,Skinderowicz2020}.  More
computational power allowed not only to reduce the computation time but also to
tackle larger problems~\cite{Chitty2016, Peake2019}.

In this paper, building upon novel ideas and recent research on applying the
ACO to large TSP instances, we propose a new algorithm loosely based on the
MMAS. The algorithm, named the Focused ACO (FACO), can be used to find high-quality
solutions (within 1\% from the best-known results) to the TSP instances with
more than one hundred thousand nodes while also being competitive in terms of
the computation time. Our straightforward parallel implementation of the algorithm
requires less than an hour to solve (approximately) the TSP instances from the
\emph{TSP Art Instances} dataset~\cite{ArtTSP} with $10^5$ (100k) to $2\cdot
10^5$ (200k) cities when executing on a computer with an 8-core commodity CPU.
The main contributions presented in this paper can be summarized as follows:
\begin{itemize}
\item
    We present a novel ACO algorithm named the FACO, which contains a mechanism
    for controlling the differences between the newly constructed solutions
    and a selected previous solution. The resulting search process is more
    narrow (focused), allowing to find improvements to small parts of the
    existing solution without worsening the quality of the rest.
\item
    We show how the control over the new components (differences) in the
    constructed solutions can be used to guide the local search~(LS)
    heuristic (2-opt) in order to reduce its computation time.
\item
    We show how the proposed ideas can be applied along the recent advancements
    presented in the literature on applying the ACO to tackle large TSP
    instances efficiently.
\item
    \R{
        We present a sensitivity analysis of the proposed FACO algorithm
        considering both the quality of the results and the computation time.
        In particular, we show how the number of new components in the
        constructed solutions affects the efficiency of the algorithm.
    }
\item
    \R{
        We describe our parallel implementation of the FACO for shared-memory
        multiprocessing. The implementation allows to efficiently utilize the
        processing power of a multicore CPU.
    }
\item
    We present an extensive computational study of the proposed algorithm and
    compare its efficiency to that of the recently proposed ACOs, and also
    with the powerful LKH solver by Helsgaun~\cite{Helsgaun2000} on the TSP
    instances with up to two hundred thousand cities (nodes).
\end{itemize}

The remainder of this paper is organized as follows. In
Sec.~\ref{sec:background} we describe the MMAS, which is the basis for the
proposed FACO, and highlight the main performance obstacles of the ACOs, including
the MMAS. Section~\ref{sec:Related_work} contains a summary of the recent research
focused on improving the performance of the ACOs. In
Sec.~\ref{sec:Adapting-ACO-to-Large-Problem-Instances} we detail the proposed
FACO, emphasizing features that are essential for the good performance when solving
large TSP instances. Section~\ref{sec:Experimental-analysis} summarizes the
experimental study of the FACO, which comprised a sensitivity analysis of its
main parameters and a comparison with the state-of-the-art ACOs. Finally,
Sec.~\ref{sec:Conclusions} summarizes the work.

\section{Background}
\label{sec:background}

\subsection{Ant Colony Optimization}
\label{sec:ACO} 

The ACO belongs to a group of swarm-based metaheuristics (SBMs) in which a number of simple information-processing units (agents) construct solutions to an optimization problem. The agents can often interact with each other, often indirectly, to improve the problem-solving efficiency~\cite{Kennedy2006}. Typically, the SBMs mimic some biological systems, including ant colonies, flocks of birds, and schools of fish, among others~\cite{Engelbrecht2005}.

The TSP was among the first combinatorial optimization problems to which the
ACO was applied~\cite{Dorigo1996}, and still to date remains a default choice
when describing and verifying the efficiency of new variants of the
ACO~\cite{Chitty2016, Ismkhan2017, Stutzle2000}.  Typically, the TSP is modeled
using a complete graph $G = (V, E)$ in which a set of nodes $V$ corresponds to
a set of cities or locations on a map, and a set of edges $E$ represents direct
connections or roads between pairs of the nodes. For convenience, the nodes can
be numbered from 0 to $n-1$ ($n$ being the number of nodes), and the set of
edges (arcs) can be defined as $E = \{ (i,j): i, j \in V, i \ne j \}$, i.e.,
each edge is represented using a pair of nodes. 
The TSP typically corresponds to a \emph{symmetric} variant in which the order
of the edges' nodes is not important. Complementary, if the edges are
\emph{directed}, then the problem is \emph{asymmetric} (ATSP). With every edge,
$(i,j)$, a positive cost, $d_{ij}$, is associated, allowing to model real-world
distances or travel costs between cities $i$ and $j$. The ATSP variant allows
the costs to differ depending on the direction of travel.  Solving the (A)TSP
requires finding a minimum cost route visiting each city exactly once, or, in
other words, the minimum cost Hamiltonian cycle in graph $G$.

In the ACO, a virtual ant travels through the graph $G$ moving from a current
node to one of the neighbor nodes. The route of the ant becomes its solution.
The decision of which node to choose next is based on the costs of the edges
leading to yet unvisited nodes and additional information in the form of
so-called \emph{pheromone trails} deposited on the edges.  Specifically, for
every edge $(i, j) \in E$ a pheromone trail, $\tau_{ij}(t)$, is defined,
where $t$ denotes discrete time. 
The use of the artificial pheromone trails mimics how some species of ants use
chemical substances as a medium of indirect communication between the
individuals~\cite{Dorigo2004}. The values (concentrations) of the artificial
pheromone trails are stored in the computer's memory as a real-valued
matrix of size $O(n^2)$. The matrix is often referred to as a \emph{pheromone
memory}.  Some variants of the ACO, including the MMAS, impose bounds on the
pheromone values, i.e., $\tau_{\rm min}$ and $\tau_{\rm max}$, which gives more
control over the exploration--exploitation behavior induced by the pheromone 

Focusing on the MMAS, which is the basis of the proposed FACO algorithm,
the solution construction process of the ACOs can be explained in
more detail as follows. Ant $k$ located at node $i$ selects edge $(i, j)$
with the probability defined as:
\begin{equation}
\label{eq:prob}
p_{ij}^k(t) = \frac{ [ \tau_{ij}(t) ]^\alpha [ \eta_{ij} ]^\beta }{ \sum_{l \in \mathcal{N}_i^k} [\tau_{il}(t)]^\alpha [ \eta_{il} ]^\beta }  \; \quad \textrm{if} \; j \in \mathcal{N}_i^k \, ,
\end{equation}\\
where $\tau_{ij}(t)$ is the value of the pheromone trail deposited on edge
$(i,j)$; $\eta_{ij}$ is the value of so-called \emph{heuristic information} of
edge $(i, j)$; $\alpha$ and $\beta$ are parameters that control the
relative influence of the pheromone values and the heuristic information,
respectively. Finally, $\mathcal{N}_i^k$, is a set of nodes (that
neighbor $i$) to be visited by ant $k$. In the case of the TSP, the value of
the heuristic information of edge $(i, j)$ is defined as a reciprocal of
the edge length (cost), i.e., $\frac{1}{d_{ij}}$, what makes the edge more
attractive the shorter it is.

In the case of the TSP, the heuristic information remains static as the
distances between the cities do not change. However, the pheromone memory is
\emph{dynamic}, i.e., the amount of pheromone can diminish due to a process
called \emph{evaporation}, or can increase to make it more likely for an edge
to be selected by the ants in subsequent iterations. Increasing the amount of
pheromone aims at inducing a positive feedback loop resulting in new solutions
similar to the best solutions found so far but with a lower cost. In contrast,
evaporating the pheromone increases the chance of building solutions with more
differences and possibly escaping from a local minimum in the solution search
space.  In other words, the pheromone deposition and evaporation allow to
balance between \emph{exploitation} and \emph{exploration},
respectively~\cite{Dorigo1996,Stutzle2000}.

In the MMAS, the pheromone values are initialized to a maximum value,
$\tau_{\rm max}(0)$. Next, the main loop of the algorithm starts. Firstly, all
ants construct their solutions to the problem. Next, the evaporation process
lowers the values of all pheromone trails according to a predefined parameter
$\rho$ but not below $\tau_{\rm min}(t)$, where $t$ denotes the current
iteration. Finally, the pheromone deposition adds a small amount of pheromone
to the edges belonging to a \emph{high quality} solution, which is either the
best solution found in the current iteration (iteration best solution) or the
best solution found to date (global best solution). Precisely, the amount of
pheromone on edge $(i,j)$ changes between the current, $t$, and next,
$t+1$, iteration according to equation:
\begin{equation}
\label{eq:pher-update}
    \tau_{ij}(t + 1) = \textrm{max} \left \{ \tau_{\rm min}(t+1),
    \textrm{min} \left \{ \tau_{\rm max}(t+1), \rho \tau_{ij}(t) + \Delta \tau_{ij}(t)
    \right \} \right \} \, ,
\end{equation}
where $\rho \in {[} 0, 1 {]}$ is a parameter controlling the amount of the pheromone that remains after the evaporation and $\Delta \tau_{ij}(t)$ is the amount of pheromone deposited on the edge $(i,j)$.

Finding a new global best solution at iteration $t$ results in adjusting the
pheromone limits, $\tau_{\rm min}(t+1)$ and $\tau_{\rm max}(t+1)$, which are
used in the subsequent iteration, $t+1$, otherwise they remain the same. The
whole process repeats for a specified number of iterations, or until another
stopping criterion is reached. Figure~\ref{alg:mmas} shows the pseudocode of the
MMAS. Time complexity of the solution construction loop
(lines~\ref{alg:mmas.build.loop.beg}--\ref{alg:mmas.build.loop.end}) equals
$O(n^2)$ as the decision of which node to choose next
($\textrm{select\_next\_node()}$ procedure) is called $n-1$ times, where $n$ is
the size of the problem, and each call takes time proportional to $n$.
If $\#\textit{ants} = n$, then a single iteration of the algorithm has
complexity of $O(n^3)$. However, if a \emph{local search} procedure is called
for every solution (line~\ref{alg:mmas.ls}), its time complexity may dominate
over the solution construction time.

\begin{algorithm}[h]
\SetKwInOut{Input}{Input}\SetKwInOut{Output}{Output}

\def\ant[#1]{\textrm{Ant}(#1)}
\def\route[#1]{\textit{route}_{\textrm{Ant}(#1)}}
\def\iterbest{\textit{iter\_best}}
\def\globalbest{\textit{global\_best}}
\def\NewEdges{\textit{new\_edges}}
\def\It{\textit{it}}
\def\Pred{\textrm{pred}}
\def\Succ{\textrm{succ}}
\def\LSChecklist{\textit{LS\_checklist}}
\def\SourceSol{\textit{source\_solution}}
\def\:={\leftarrow}

$\globalbest \:= \textrm{Build initial solution}$ \label{alg:mmas.init.sol}

Calculate pheromone trails limits $\tau_{\rm{min}}$ and
$\tau_{\rm{max}}$ using $\globalbest$ \\

Set pheromone trails values to $\tau_{\rm{max}}$ \label{alg:mmas.pherinit} \\

\For{ $i \:= 1$ \KwTo $\textit{\#iterations} $ }{
    \label{alg:mmas.main.loop.start}
    
    \For{ $j \:= 0$ \KwTo $\textit{\#ants}-1$ }{
        $\route[j] \left[0\right] \:= \mathcal{U}\{0, n-1\}$ \quad \tcp{Select first node randomly}

        \For{ $k \:= 1$ \KwTo $n-1$ }{  \label{alg:mmas.build.loop.beg}
            $\route[j] \left[k\right] \:= \textrm{select\_next\_node}( \route[j] )$ \\

        } \label{alg:mmas.build.loop.end}

        $\textrm{local\_search}(\route[j])$  \label{alg:mmas.ls} \quad \tcp{Optional}
    }
    
    $\iterbest \leftarrow \textrm{select\_shortest} \left(\route[0], \ldots, \route[\#ants-1] \right)$ 
    \label{alg:mmas.iter.best} \\
    
    \If{ $\iterbest$ {\rm is shorter than} $\globalbest$ }{
        $\globalbest \leftarrow \iterbest$ \\
        Update pheromone trails limits
        $\tau_{\rm{min}}$ and $\tau_{\rm{max}}$ using $\globalbest$
    }

    Evaporate pheromone according to $\rho$ parameter \label{alg:mmas.pher.evap}

    Deposit pheromone using either $\globalbest \textrm{ or } \iterbest$ solution
    \label{alg:mmas.pher.deposit}

   \label{alg:mmas:main.loop.end}
}
\caption{The MAX-MIN Ant System.}
\label{alg:mmas}
\end{algorithm}


\subsection{ACO Performance Obstacles}
\label{sec:perf-obstacles}

In the basic form, the ACO algorithms are capable of solving only very
small TSP instances with dozens of cities~\cite{Dorigo2004, Stutzle2000}.
There are three main components of the ACO that affect its performance in terms
of the execution time and quality of the generated solutions.  The first is the
(next) node selection procedure used by the ants during the solution
construction process. The second is the pheromone memory which can be expensive
to store. The third component is the LS heuristic applied to improve the
quality of the solutions constructed by the ants.  A significant part of the
ACO-related research known from the literature addresses at least one of the
components mentioned. 
The proposed ideas include optimized implementations of the selected ACO
components, employ parallel computations~\cite{Pedemonte2011}, and, finally,
change the inner workings of the components.  

\section{Related work}
\label{sec:Related_work}

In this section, we briefly summarize the recent research on the ACO, focusing
mainly on the computing efficiency and speed of convergence to solutions of
good quality, which are essential if one considers solving TSP instances with
dozens or even hundreds of thousands of nodes.

\subsection{Next Node Selection Improvements}
\label{sec:Next-Node-Sel-Impr} 

One of the first modifications to the Ant System by~Dorigo et
al.~\cite{Dorigo1996} attempts to speed up the solution construction process by
limiting the number of choices considered when calculating probabilities as
defined by~Eq.~(\ref{eq:prob})~\cite{Dorigo1997}.  Only the edges connecting the current node
with a number of its closest neighbors, so-called \emph{candidate list}, are
considered. If all of the nearest neighbors of the current node
were visited, one of the remaining nodes is selected instead. The typical choice is the
node to which leads an edge with the highest product of the pheromone
concentration and heuristic information value.
The candidate lists significantly reduce the computation time while still
allowing the algorithm to find high-quality solutions. The explanation comes
from an observation that in many real-world TSP instances, the optimal
solutions consist primarily of short edges connecting nodes that are close
neighbors~\cite{Helsgaun2000}. Typically the length of the candidate lists,
$\textit{cl\_size}$, is a small constant, e.g., 10 to 30, allowing to reduce
the computational complexity of the node selection from $O(n)$ to $O(1)$ if the
respective candidate list contains at least one unvisited node.

Recently Mart{\'{\i}}nez and~Garc{\'{\i}}a have introduced a similar idea of
\emph{backup cities} or \emph{backup lists}, which contain a (constant) number
of nodes which are the first of the nearest neighbors that did not fit into the
respective candidate lists~\cite{Martinez2021}.  The backup lists are used if
all nodes belonging to the candidate list of the current node have already been
visited (are part of the constructed solution).  The authors considered two
usages for the backup lists, namely \emph{conservative} and \emph{aggressive}. 
In the conservative mode, the computations are the same as in the case of the
candidate lists, i.e., the probability of selecting a node follows
Eq.~(\ref{eq:prob}). In the aggressive mode, no pheromone is used for the edges
connecting the current node with the nodes from the backup list, and the
selection procedure reduces to picking the closest yet unvisited node from the
list. The authors have found that the aggressive mode leads to a faster execution
of the proposed ACOTSP-MF without sacrificing the quality of the solutions
produced when solving the TSP instances with up to 200k (or $2 \cdot 10^5$)
cities.

Another interesting idea related to the candidate lists was proposed by~Ismkhan
in his Effective Strategies+ACO (ESACO) algorithm, which demonstrated
a state-of-the-art performance (among ACO-based approaches) when solving the TSP
instances of size up to \numprint{18512} cities~\cite{Ismkhan2017}. In the ESACO, the candidate
lists are \emph{dynamic} and updated based on the best solution found so far,
so that if edge $(u, v)$ is a part of the best solution, then $v$ is
inserted at the beginning of the candidate list of the node $u$. This idea is
beneficial if node $v$ does not fit into the static candidate list. The results
confirmed a modest improvement in the quality for some of the TSP instances
considered.

Recent research includes also a few ideas concentrated on the computing
efficiency of the node selection. For example, modern CPUs offer rich sets of
SIMD instructions for speeding up computations, but often, the implementation
has to be adapted to benefit from them~\cite{Zhou2018, Martinez2021}.
Analogously, modern GPUs can efficiently execute the ACO-based algorithms,
including the ACS and MMAS, but the implementation has to consider the
specifics of the GPU architecture~\cite{Cecilia2018, Skinderowicz2016,
Skinderowicz2020}.

Although the candidate lists and careful, performance-oriented optimizations
are effective, the solution construction process can be sped up even further by
introducing more determinism during the next node selection process.
Gambardella et al.~\cite{Gambardella2012} proposed the Enhanced ACS (EACS)
algorithm in which an ant located at node $u$ selects with a high probability
the node which follows $u$ in the best solution computed so far. In other
words, most of the edges of the constructed solution will be copied from the
current best solution, avoiding more time-consuming computations of
probabilities based on the pheromone memory and heuristic information. Coupled
with efficient local search, the EACS outperformed the ACS when solving the
sequential ordering problem, the probabilistic TSP, and the TSP.

The idea of utilizing the existing solutions to speed up the construction
process of new solutions was also noticed by~Chitty~\cite{Chitty2016}, who
developed the PartialACO algorithm aimed at solving TSP instances with dozens
to hundreds of thousands of nodes. In the PartialACO, an ant starts the
solution construction process by copying a part of a \emph{local} best
solution, which is the best solution found so far by the ant. Finally, the
remaining part of the route is completed probabilistically as in the
Population-based ACO by~Guntsch~\cite{Guntsch2002}. Not surprisingly, limiting
the number of choices an ant has to make leads to a significant speed
improvement, proportional to the length of the copied part. The algorithm was able to find
solutions within a few percent from the best-known results for the TSP
instances with up to 200k nodes when executing on a computer with 8-core Intel
Core i7 CPU~\cite{Chitty2016}.

\subsection{Reducing Pheromone Memory Size}
\label{sec:red-pher-mem-size} 

The ACO algorithms, including the ACS and MMAS, have relatively high memory
complexity of $O(n^2)$, where $n$ is the size of the problem, resulting mainly
from the pheromone memory storage. If the value of a pheromone is stored for
each edge connecting the cities in the TSP, then, typically, a matrix of size
$n \times n$ is used as a data structure, often referred to as the
\emph{pheromone matrix}. ACO implementations often employ other matrices that
play the role of caches. For instance, it is convenient to store a matrix
of size $n \times n$ of (precomputed) distances between the cities. The
distance calculations often require costly instructions, and accessing the
pre-calculated data stored in the matrix can be more efficient, especially if
the matrix fits in the computer's cache memory. Analogously, the products of the pheromone
values and heuristic information, necessary to calculate the probabilities
described by Eq.~(\ref{eq:prob}), can also be stored in a separate
matrix~\cite{Dorigo2004}.
Unfortunately, if one considers solving large TSP instances, then storing these
data structures in the memory becomes a problem, especially when the GPUs are
considered~\cite{Cecilia2018, Skinderowicz2020}.  For example, if $n = 10^5$
and 32-bit floating-point numbers are used to store the pheromone values, then
the pheromone matrix alone takes over 37GB of memory. Even if enough RAM is
available, reads and writes to the memory matrix may still hinder the
performance due to relatively slow memory access times in modern
systems~\cite{Martinez2021}. Peake et al.~\cite{Peake2019} proposed a
Restricted Pheromone Matrix in which the current pheromone values are stored
only for the edges connecting a node to its nearest neighbors in the
corresponding candidate list. The length of the candidate lists is typically a
small constant, resulting in the memory complexity of the pheromone memory
being reduced to $O(n)$. 
A close idea was proposed in our earlier work~\cite{Skinderowicz2013}
introducing a \emph{selective pheromone memory} in which only a small, constant number of
edge--pheromone pairs were stored for every node. However, the sets of edges
were not restricted to the candidate lists and could change based on the access
patterns, following well-established ideas of cache memory implementations.
Ismkhan also proposed a similar idea of restricting the pheromone memory size but
allowing dynamic changes and applied it in his ESACO
algorithm~\cite{Ismkhan2017}.

Probably the most radical idea was proposed by~Guntsch and~Middendorf, who
decided to remove pheromone memory entirely in their Population-based ACO
(PACO)~\cite{Guntsch2002}. In the PACO,  a \emph{population} of solutions
chosen among the previously constructed solutions replaces the explicit
pheromone memory. The pheromone concentration for a given edge is computed as
needed based on the current contents of the population.  The PACO was an
inspiration for Chitty, who introduced the PartialACO in which an explicit
pheromone matrix is replaced with a population of solutions.  However, the
population is split among the ants so that each ant stores its best solution
found so far~\cite{Chitty2016}.

While reducing the pheromone memory size is undoubtedly challenging, reducing
the size of the remaining data structures used in the ACO algorithms is
somewhat easier. For example, the distances between cities can be calculated on
demand, removing the necessity for storing the distance matrix. Alternatively,
the precomputed distances can be stored only for the nearest neighbors of each
node~\cite{Martinez2021}.  Overall, it is possible to reduce the memory
complexity of the ACO by an order of magnitude, i.e., from $O(n^2)$ to $O(n)$.

\subsection{Local Search}
\label{sec:local-search} 

The ACO-based algorithms are typically paired with an efficient LS 
heuristic which is essential for obtaining high-quality solutions,
especially if the size of the TSP instances is on
the order of thousands or more~\cite{Dorigo2004, Pedemonte2011, Stutzle2000}.
When combined with the LS, the ACO is responsible for making "jumps" between
possibly distant regions of the solution search space, while the LS is
responsible for locating local optima by improving the solutions generated by
the ants. The LS can be applied to all or a subset of the solutions,
periodically or with a specified probability.
One of the simplest LS heuristics for the TSP is the 2-opt heuristic which
searches for a pair of disjoint edges that can be replaced with a new pair of
edges but of a smaller total length (cost). As the 2-opt move is equivalent to
reversing a route section, there are $O(n^2)$ possible moves to consider.
Fortunately, the 2-opt can be significantly sped up if only a small subset of
all possible pairs is considered~\cite{Bentley1992}.  The 2-opt can be
generalized to $k$-opt form, which exchanges $k$ edges at a time, where $k >
2$. Often, the $k$-opt ($k \ge 3$) moves can improve the quality of the
solutions (relative to $2$-opt) at the cost of a significantly increased
computation time~\cite{Helsgaun2009}.

The constructive heuristics, including the ACO, produce a large number of
solutions to the tackled problem, so the time complexity of any LS heuristic
used should be relatively small~\cite{Dorigo1996, Stutzle2000}.  The ACO
algorithms (in the context of the TSP and related problems) often use
relatively fast heuristics, including the 2-opt and 3-opt;  even then, the LS
can be applied to only a subset of the constructed solutions~\cite{Chitty2016,
Martinez2021}. If the number of solutions constructed is small, it is possible
to apply more complex LS heuristics as in the ESACO proposed
by~Ismkhan~\cite{Ismkhan2017} in which the LS considers 2-opt, 3-opt, and even
some 4-opt (double-bridge) moves.


\subsection{Other Metaheuristics for Solving Large TSP Instances}

The ACOs are not the only metaheuristics applied to solving the TSP.  The
literature contains numerous examples of both exact and approximate approaches
to solving the problem. For the sake of clarity, we mention here some of the
most effective (in terms of the produced solutions quality) works on applying
nature-inspired metaheuristics to solving large TSP instances.

Marinakis et al. ~\cite{Marinakis2010} proposed an efficient Honey Bees Mating
Optimization Algorithm for the Traveling Salesman Problem (HBMOTSP), which
combines several methods, including the Honey Bees Mating Optimization
algorithm, the Multiple Phase Neighborhood Search-Greedy Randomized Adaptive
Search Procedure, and the Expanding Neighborhood Search Strategy with the
2-opt, 2.5-opt, and the 3-opt TSP heuristics proposed by~Lin~\cite{Lin1965}.
The HBMOTSP was able to solve all the TSP instances from the TSPLIB (with up to
\numprint{85900} nodes), achieving an average relative error below 1\%.
Zhong et. al~\cite{Zhong2018} proposed the Discrete Pigeon-inspired
Optimization (DPIO) in which a flock of virtual pigeons moves through the
solution search space. Positions of the pigeons represent valid solutions to
the TSP, and new positions are calculated by so called flying operators
resulting in perturbation of the current solutions (positions). Combined with
the Metropolis acceptance criterion known from the Simulated Annealing, the
resulting method produced solutions to the largest TSP instances from TSPLIB
repository with an average error less than 2\% relative to the optima.
Choong et al.~\cite{Choong2019} proposed the Artificial Bee Colony algorithm with a Modified
Choice Function (MFC-ABC), another example of nature-inspired, perturbation
metaheuristic. The MFC-ABC combined with the well known
Lin-Kernighan~\cite{Lin1973} heuristic was able to find good quality solutions
(below 1\% from the optima) to all of the TSPLIB instances.

Unsurprisingly, the most successful approaches are examples of the
perturbation-based algorithms, as modifying an existing solution is, typically,
faster than constructing a new one from scratch, while also preventing large
"jumps" in the solution search space. Additionally, efficient LS heuristics
play a significant role in improving the overall quality of the produced
solutions, especially for large TSP instances.

\section{\R{Focused ACO for Solving Large Problem Instances}}
\label{sec:Adapting-ACO-to-Large-Problem-Instances}

\R{
    This section describes the proposed FACO algorithm in detail, while the
    following section contains a related computational study involving TSP
    instances with up to 200k nodes.  Scaling the ACO-based algorithm to solve
    large problem instances requires several changes to its core components to
    reduce the memory and computational complexities.
    The proposed FACO is a careful combination of novel ideas (solution
    construction process with precisely controlled exploration, LS checklists),
    with proven solutions from the literature (partial pheromone memory,
    candidate, and backup lists) wrapped in an efficient parallel
    implementation. Additionally, the efficiency of the FACO's search process
    depends strongly on the values of its parameters as described in
    Sec.~\ref{sec:pheromone-trail-limits} and
    Sec.~\ref{sec:Experimental-analysis}.
}


\subsection{\R{Candidate and Backup Lists}}

\R{
The proposed FACO applies the idea of \emph{candidate lists} that
significantly sped up the solution construction process. Specifically, the
candidate lists limit the number of choices an ant can make only to a small
subset of the \emph{nearest neighbors} of each node. 
}
Recently Mart{\'{\i}}nez and
Garc{\'{\i}}a have shown how to further improve the execution speed with the
help of so-called \emph{backup lists}~\cite{Martinez2021}. The lists contain a
specified number of the next to the nearest neighbors of every node, i.e., the
first nodes that did not fit into the candidate list and we also apply this
idea.

\subsection{\R{Partial Pheromone Memory}}

\R{
As the size of the problem increases to tens of thousands of nodes, the size
of the pheromone memory, which is quadratic, becomes problematic,
especially on systems with a smaller amount of RAM, including dedicated
accelerators like the GPUs~\cite{Cecilia2018, Skinderowicz2016}.  And even on
systems with sufficient RAM, the smaller (partial) pheromone memory is
preferred as the relative size of the part that fits into (high-speed) CPU
caches increases.
}
One solution to the problem is to
store the values of the pheromone only for a small subset of all possible
solution components (edges in the case of the TSP)~\cite{Ismkhan2017, Skinderowicz2013,
Skinderowicz2016}, while the other is to not store the pheromone
values at all~\cite{Chitty2016, Guntsch2002}.  In our work, we follow the first
approach and store the pheromone values only for the edges which connect a
given node to a number of its closest neighbors. Specifically, by the closest
neighbors, we mean the nodes that belong to a node's candidate list. As the
size of the candidate list is a small constant, this reduces the pheromone
memory complexity from $O(n^2)$ to $O(n)$.


\subsection{Solution Construction Process with Limited Exploration}
\label{sec:Solution_Construction_Process_with_Limited_Exploration}

The solution construction process in the ACO algorithms allows for some degree
of freedom, i.e., the probability of selecting a given component (edge) is
defined by Eq.~\ref{eq:prob}, and in effect, it depends on the amount of the
pheromone deposited. Even if the highest possible amount of pheromone belongs
only to the \emph{best edges}, i.e., the edges of the best solution found so
far, the other edges still have a non-zero probability of being selected. The
pheromone values have to be positive for this to hold, as in the ACS and MMAS.
Although this property is essential for allowing the search process to progress
and escape local minima, it can also lead to slow convergence,
especially if the size of the problem is large.  The problem is that the number
of times the construction process selects a not-best edge grows with the
problem size. 
For example, let us assume that the probability of selecting an edge with the
highest pheromone always equals 99.9\%. The probability that the constructed
solution will consist of only the edges with the highest pheromone
concentration is proportional to $0.999^{n-1}$, where $n$ is the number of
nodes. This is close to $90\%$ if $n=100$ but for $n=1000$ it drops close to
$37\%$, and for $n=10000$ it becomes close to 0. 

Another important issue is that the new solution components are most likely to
be spread along the constructed tour. This can be less beneficial than focusing
on a few smaller parts of the solution, especially for large
problems~\cite{Taillard2018}. Intuitively, improving a specific part of a
solution may require introducing several changes (new edges) relatively close
to each other. Following this line of reasoning, we propose a modification of
the ACO solution construction process addressing both the number of new edges
and their localization within a route. Specifically, we propose to start the
solution construction from a randomly selected node, then executing the ACO
choice rule until a \emph{specified} number of \emph{new edges} are selected,
i.e., edges that are not present in a \emph{source solution}. The source
solution can be any of the solutions built so far -- in our algorithm, it is
selected probabilistically between the iteration best and global best
solutions.  When enough new (different) edges get selected, the solution is
completed by copying edges from the source solution.

\begin{algorithm}[H]
\SetKwInOut{Input}{Input}\SetKwInOut{Output}{Output}

\def\ant[#1]{\textrm{Ant}(#1)}
\def\route[#1]{\textit{route}_{\textrm{Ant}(#1)}}
\def\iterbest{\textit{iter\_best}}
\def\globalbest{\textit{global\_best}}
\def\NewEdges{\textit{new\_edges}}
\def\It{\textit{it}}
\def\Pred{\textrm{pred}}
\def\Succ{\textrm{succ}}
\def\LSChecklist{\textit{LS\_checklist}}
\def\SourceSol{\textit{source\_solution}}
\def\:={\leftarrow}

$\globalbest \:= \textrm{Build initial solution}$ \label{alg:faco.init.sol}

Calculate pheromone trails limits: $\tau_{\rm{min}}$ and
$\tau_{\rm{max}}$ \\

Set pheromone trails values to $\tau_{\rm{max}}$ \label{alg:faco.pherinit} \\

$\SourceSol \:= \globalbest$ \label{alg:faco.source.init}

\For{ $i \:= 1$ \KwTo $\textit{\#iterations} $ }{
    \label{alg:faco.main.loop.start}
    
    \For{ $j \:= 0$ \KwTo $\textit{\#ants}-1$ }{
        \label{alg:faco.ants.loop.start}
        $\route[j] \left[0\right] \:= \mathcal{U}\{0, n-1\}$ \quad \tcp{Select first node randomly}

        $\textit{min\_new\_edges} \:= \textrm{calc\_num\_new\_edges()}$ \\
        
        $\textit{new\_edges} \:= 0$ \\
        
        $k \:= 1$ \\

        \While{ $k < n$ }{  \label{alg:faco.build.loop.beg}

            $u \:= \route[j] \left[k-1\right] $ \\

            $v \:= \textrm{select\_next\_node}( u, \route[j] )$ \\

            $\route[j] \left[k\right] \:= v$ \\
            $k \:= k + 1$ \\

            \If{  $(u, v) \notin \SourceSol $ }{ \label{alg:faco.new.edge.check}

                $\textit{new\_edges} \:= \textit{new\_edges} + 1$ \\

                Add $v$ to $\LSChecklist$ \\
            }

            \If{ $\NewEdges \ge \textit{min\_new\_edges}$ }{
                \tcp{Complete $\route[j]$ following $\SourceSol$\ldots}
                \label{alg:faco.complete.beg}

                $ u \:= \Succ( \SourceSol, v )$  \quad \tcp{\ldots forward\ldots}

                \While{ $u \notin \route[j]$ }{  
                    $\route[j] \left[k\right] \:= u$

                    $ u \:= \Succ( \SourceSol, u )$

                    $k \:= k + 1$ \\
                }

                $u \:= \Pred( \SourceSol, u )$   \quad \tcp{\ldots or backward}

                \While{ $u \notin \route[j]$ }{
                    $\route[j] \left[k\right] \:= u$

                    $ u \:= \Pred( \SourceSol, u )$

                    $k \:= k + 1$
                }
                \label{alg:faco.complete.end}
            }
        } \label{alg:faco.build.loop.end}

        $\textrm{local\_search}(\route[j], \LSChecklist)$  \label{alg:faco.ls}
    }
    
    $\iterbest \leftarrow \textrm{select\_shortest} \left(\route[0], \ldots, \route[\#ants-1] \right)$ 
    \label{alg:faco.iter.best} \\
    
    \If{ $\globalbest = \emptyset$ {\bf or} $\iterbest$ {\rm is shorter than} $\globalbest$ }{
        $\globalbest \leftarrow \iterbest$ \\
        Update pheromone trails limits
        $\tau_{\rm{min}}$ and $\tau_{\rm{max}}$ using $\globalbest$
    }

    Evaporate pheromone according to $\rho$ parameter \label{alg:faco.pher.evap}

    $\SourceSol \:= \textrm{Choose between } \globalbest \textrm{ and } \iterbest$

    Deposit pheromone \label{alg:faco.pher.deposit}

   \label{alg:faco:main.loop.end}
}
\caption{The Focused Ant Colony Optimization.}
\label{alg:faco}
\end{algorithm}

Figure~\ref{alg:faco} shows the pseudocode of the FACO.  The algorithm starts
with the construction of an initial solution (line~\ref{alg:faco.init.sol}),
which in our implementation is created using the nearest neighbor heuristic and
further improved with the 2-opt and 3-opt LS the implementations following
Bentley~\cite{Bentley1992}.  The initial solution becomes the first \emph{source
solution} (line~\ref{alg:faco.source.init}).  In the main loop of the algorithm
(lines~\ref{alg:faco.main.loop.start}--\ref{alg:faco:main.loop.end}), each ant
constructs a solution to the problem starting at a randomly selected node.
Next, the ant iteratively travels to one of the (yet unvisited) neighbor nodes.
Equation~(\ref{eq:prob}) defines the probability of selecting an unvisited node
from the \emph{candidate list} corresponding to the current node. If the ant
has already visited all nodes on the list, then it chooses the first unvisited
node from the respective \emph{backup list}, following the idea by~Martinez and
Garcia~\cite{Martinez2021}. Finally, if no such node exists, the ant selects a
node corresponding to the edge with the highest value of heuristic information.
In the case of the TSP, this is equivalent to selecting the shortest edge.

After deciding on the next node, we check (line~\ref{alg:faco.new.edge.check} in
Fig.~\ref{alg:faco}) whether the corresponding edge is \emph{present} in the
source solution. If not, then the counter of \emph{new edges} is
increased.  If enough new edges are already in the constructed solution
($\textit{new\_edges} \ge \textit{min\_new\_edges}$) then the remaining edges
are \emph{copied}
(lines~\ref{alg:faco.complete.beg}--\ref{alg:faco.complete.end} in
Fig.~\ref{alg:faco}) from the source solution.
It is worth mentioning that the actual number of new edges introduced in the
constructed solution may be higher as we do not count the last edge, which
closes the tour. Furthermore, the new edges included in the constructed
solution may prevent copying some edges from the source solution. For example,
if a pair of (new) edges, $(a, b)$ and $(b, c)$, have been added to the
constructed solution, then it is not possible to copy an edge $(c, a)$ from the
source solution as this would introduce a cycle.

\subsection{\R{Speeding up the Local Search with Checklists}}

After the construction process finishes, the 2-opt LS heuristic
(line~\ref{alg:faco.ls} in Fig.~\ref{alg:faco}) tries to improve the new
solution. The construction process keeps track of the differences (new edges)
between the new and the source solutions, allowing to pass the list of nodes,
\emph{LSChecklist}, to check for an improving move (change). In other words,
the LS \emph{skips} checking the parts of the solution that are the same as in the
source solution. If the length of the \emph{LSChecklist} is small, then this
approach can provide a significant speedup over the full check of all nodes.

Figure~\ref{alg:twoopt} shows the pseudocode of the 2-opt LS procedure. The inputs are
the route to improve and the list, \emph{LSChecklist}, of nodes to check for 
improving moves (changes). A valid move consists of a pair of edges. The first
edge of the pair contains as one of its endpoints node $a$, 
$a \in \textit{LSChecklist}$, while the other edge contains node $b$ such that
$b$ is one of the nearest neighbors of node $a$. Limiting the search to only
the nearest neighbors of the considered node is a common optimization that
reduces the computation time significantly, as the number of the nearest
neighbors is typically a small constant~\cite{Bentley1992}. If a valid move is found, it
is applied by flipping the corresponding route section, and the endpoints of
the move's edges are added to \emph{LSChecklist}. The algorithm continues until
\emph{LSChecklist} becomes empty or a limit of valid changes is reached. The
time complexity of the whole procedure is $O(kn)$, where $k$ is the number of
improving moves found, and $n$ denotes the size of the problem. 
As $k \le n$ the complexity of the LS in the worst case is $O(n^2)$.

\subsection{Note on Pheromone Trail Limits}
\label{sec:pheromone-trail-limits} 

In the MMAS the pheromone deposited on the solution components (edges) belongs
to a range ${[} \tau_{\rm min}, \tau_{\rm max} {]}$~\cite{Stutzle2000}.  The
maximum value is set to $\frac{1}{{\it cost}_{\rm gb} (1 - \rho)}$, where ${\it
cost}_{\rm gb}$ is the cost of the best solution found so far and $\rho$
determines how much of the pheromone is retained between successive iterations.
Setting the minimum pheromone value, $\tau_{\rm min}$, to a non-zero value
assures that each available solution component has a positive probability of
being selected (see~Eq.~(\ref{eq:prob})). 
St{\"{u}}tzle and Hoos~\cite{Stutzle2000} proposed
the following formula
$\tau_{\rm min} = \frac{ \tau_{\rm max} (1 - \sqrt[n]{p_{\rm best}}) }{
    (\textit{avg} - 1) \sqrt[n]{ p_{\rm best} } }$, where
$n$ is the size of the problem, $p_{\rm best}$ is a parameter denoting a probability 
that the constructed solution will consist solely from the edges with the highest 
pheromone concentration, and $\textit{avg}$ is the average number of unvisited edges
considered while calculating the probabilities according to~Eq.~(\ref{eq:prob}).
In the MMAS $\textit{avg}$ is approximated by $n/2$ meaning that 
an ant chooses on average between $n/2$ solution components.
While designing the FACO we have found that a better idea 
is to set $\textit{avg}$ to $\textit{cl\_size}$, if the LS is applied. In other words,
the relative contrast between the least and the most attractive solution
components should be smaller, making the construction process more exploratory.

The proposed FACO algorithm is similar to the PartialACO by
Chitty~\cite{Chitty2016} but with a few key differences. Firstly, the FACO
controls the \emph{minimum} number of new components (edges) in the constructed
solution. In fact, the construction process follows the standard ACO approach
until the number of new (different) edges matches the specified threshold. Only
then is the solution completed by copying the remaining edges from the source
solution. On the other hand, the solution construction of the PartialACO starts
with copying a portion of one of the previous solutions. Then it completes the
solution following the standard ACO approach. In general, the completion
process does not guarantee that the selected edges will differ from the
previous solution.  In other words, the construction process of the PartialACO
has an upper bound on the number of differences relative to previous solutions.
Secondly, in the FACO, the pheromone matrix is still present and used, although the pheromone is stored only for the edges connecting a node to its (nearest) neighbors from the candidate list. Instead, the PartialACO replaces the pheromone memory with a pool of solutions following the idea of the Population-based ACO~\cite{Guntsch2002}. Finally, by keeping track of the new components (edges), the FACO allows for more efficient integration with the LS.

\begin{algorithm}[h]
\SetKwInOut{Input}{Input}\SetKwInOut{Output}{Output}

\def\ant[#1]{\textrm{Ant}(#1)}
\def\cost[#1]{\textit{cost}_{\rm #1}}
\def\route{\textit{route}}
\def\iterbest{\textit{iter\_best}}
\def\globalbest{\textit{global\_best}}
\def\NewEdges{\textit{new\_edges}}
\def\It{\textit{it}}
\def\Pred{\textrm{pred}}
\def\Succ{\textrm{succ}}
\def\Distance{\textrm{get\_distance}}
\def\LSChecklist{\textit{LS\_checklist}}
\def\SourceSol{\textit{source\_solution}}
\def\CheckQueue{\textit{LS\_checklist}}
\def\Instance{\textit{instance}}
\def\Changes{\textit{changes}}
\def\NNList{\textit{NN\_list}}
\def\Diff{\textit{diff}}
\def\gain{\textit{gain}}
\def\move{\textit{move}}
\def\:={\leftarrow}

    \Input{\route $[0,\ldots,n-1]$}
    \Input{\CheckQueue~~Starting nodes for the 2-opt moves}

    $\Changes \:= 0$ \quad \tcp{\# of successful 2-opt moves (changes) applied}

    \While{ $\CheckQueue \ne \emptyset$ \textbf{and} $\Changes < n$  }{

        $a \:= $ pop(\CheckQueue) \quad \tcp{Remove first element (node)}

        $a_{\rm succ} \:= \Succ(\route, a)$

        $a_{\rm pred} \:= \Pred(\route, a)$

        $\NNList \:= $ nearest\_neighbors(a)

        $\move \:= \emptyset$ \quad \tcp{The best 2-opt move for $\NNList$}

        $\gain \:= 0$ \quad \tcp{Cost change for $\move$}

        \ForEach{node $b \in \NNList$}{

            $b_{\rm succ} \:= \Succ(\route, b)$

            \If{ $\Distance(a, a_{\rm succ}) > \Distance(a, b)$ }{

                $\cost[old] \:= \Distance(a, a_{\rm succ}) + \Distance(b, b_{\rm succ})$

                $\cost[new] \:= \Distance(a, b) + \Distance(a_{\rm succ}, b_{\rm succ})$

                \If{ $\cost[old] - \cost[new] > \gain$ }{

                    $\gain \:= \cost[old] - \cost[new]$

                    $\move \:= (a, a_{\rm succ}, b, b_{\rm succ})$
                }

            }

        }

        \ForEach{node $b \in \NNList$}{

            $b_{\rm pred} \:= \Pred(\route, b)$

            \If{ $\Distance(a_{\rm pred}, a) > \Distance(a, b)$ }{

                $\cost[old] \:= \Distance(a_{\rm pred}, a) + \Distance(b_{\rm pred}, b)$

                $\cost[new] \:= \Distance(a, b) + \Distance(a_{\rm pred}, b_{\rm pred})$

                \If{ $\cost[old] - \cost[new] > \gain$ }{

                    $\gain \:= \cost[old] - \cost[new]$

                    $\move \:= (a_{\rm pred}, a, b_{\rm pred}, b)$
                }

            }
        }

        \If{ $\move \ne \emptyset$ }{
            $(w,x,y,z) \:= \move$

            Flip a section of $\route$ between $x$ and $y$

            Append $w, x, y, z$ to \CheckQueue

            $\Changes \:= \Changes + 1$
        }
}

\caption{The 2-opt LS used in the main loop of the FACO.}
\label{alg:twoopt}
\end{algorithm}


\R{
\subsection{Parallel Implementation}
\label{sec:Parallel-Implementation}

The multi-agent nature of the ACOs makes them a good target for parallel
execution. However, the overall effectiveness of such an approach depends to a
great extent on how the pheromone memory is used and updated by the
ants~\cite{Pedemonte2011}. In the MMAS (and the FACO), the ants share the
pheromone memory and access it in read-only mode during the solution
construction process. The only changes to the pheromone memory are applied in
subsequent steps. These involve evaporating a small amount from every trail and
deposition of additional pheromone on the trails corresponding to the best
solution found so far (alternatively, the iteration-best solution). We
implemented a straightforward parallel version of the FACO in C++ programming
language utilizing the OpenMP interface for shared-memory multiprocessing based
on these observations.
Specifically, a specified number of threads execute the solution construction
loop (Fig.~\ref{alg:faco}, lines
\ref{alg:faco.ants.loop.start}--\ref{alg:faco.ls}) in parallel. If the number
of ants exceeds the number of threads, then a single thread may do the
computations for several ants, one by one.  In our implementation, the
thread-to-ant assignment is performed dynamically in a work-stealing manner
(using \texttt{\#pragma omp for schedule(dynamic, 1)} directive).

After building the solutions, a single thread selects the iteration-best
solution and updates the global best if necessary.  Next, all threads perform
pheromone evaporation.  Finally, a single thread deposits pheromone based on
the current \emph{source solution} (line~\ref{alg:faco.pher.deposit}
in~Fig.~\ref{alg:faco}).  In our experiments with large TSP instances with at
least 100k nodes (see Sec.~\ref{sec:large-tsp-instances}), the pheromone
deposition accounted for approximately 5\% of the total computation time.
Overall, the most time-demanding operations of the FACO can be computed
efficiently in parallel resulting in close to linear speedups on our test
machine with an 8-core CPU.
}

\section{Experimental analysis}
\label{sec:Experimental-analysis}

In the following section, we investigate how the specific components and
parameters of the proposed algorithm contribute to its efficiency. Following
the sensitivity analysis, we compare the proposed FACO to the state-of-the-art
ACO-based approaches for solving the TSP. Finally, as a reference point, we
compare the results to that of the LKH solver by Helsgaun~\cite{Helsgaun2000}
which is an improved and highly optimized version of the Lin-Kernighan
heuristic, and the current state-of-the-art heuristic approach to solving TSP
instances of various sizes.

\subsection{Computing environment}
\label{sec:Computing_environment} 

The implementation of the proposed algorithms was done in C++.
Sources~\footnote{Sources are available at
https://github.com/RSkinderowicz/FocusedACO} were compiled using GCC v9.3 with
a \emph{-O3} optimization switch. The computations were conducted on a computer
with AMD Ryzen 7 4800HS 8-core CPU and 16 GB of RAM running under Ubuntu 20.04
Linux OS.  If not stated otherwise, the computations for each set of parameter
values and the TSP instance were repeated 30 times.  Also, our implementation
of the FACO was parallel (using OpenMP); hence the computations benefited from
all of the available CPU cores. On the other hand, no advanced parallel
techniques were used, e.g., explicit SIMD instructions like in the work of~Zhou
et al.~\cite{Zhou2018}.
In the comparisons concerned with execution time, parallel algorithms have an obvious advantage from the computational perspective. However, essentially all modern CPUs comprise multiple (even dozens) computing cores, and such comparisons can be seen as an answer to the question: How fast are the compared algorithms if they are executing on a single CPU.


\subsection{Focused ACO Parameters}
\label{sec:FACO_Parameters} 

The FACO inherits most of its parameters from the MMAS.  The only new parameter
is $\textit{min\_new\_edges}$ which determines the minimum number of \emph{new
edges}, i.e., edges that are in the newly constructed solution but not in the source solution.
Table~\ref{tab:parameters} shows the default values of the parameters.

\begin{table}[]
\centering
\caption{List of parameters of the proposed FACO algorithm}
\label{tab:parameters}
\begin{tabular}{@{}llrr@{}}
\toprule
\multicolumn{1}{c}{\multirow{2}{*}{Parameter}} &
  \multicolumn{1}{c}{\multirow{2}{*}{Description}} &
  \multicolumn{2}{c}{Default value(s)} \\
\multicolumn{1}{c}{} &
  \multicolumn{1}{c}{} &
  \multicolumn{1}{c}{MMAS} &
  \multicolumn{1}{c}{FACO} \\ \midrule
$m$                 & Number of ants                   & $n$                       & $ \ll n$ \\
$\rho$              & Evaporation factor               & Typically $\in [0.5, 1)$ & $(0, 1)$ \\
$\beta$             & Heuristic information importance & 2                         & 1        \\
$\textit{cl\_size}$ & Length of the candidate lists    & 16                        & 16       \\
$\textit{bl\_size}$ & Length of the backup lists       & NA                        & 64       \\
$\textit{min\_new\_edges}$ &
  \begin{tabular}[c]{@{}l@{}}Min. number of new (different) edges\\ in the constructed solution (see ?)\end{tabular} &
  NA &
  8 \\ \bottomrule
\end{tabular}
\end{table}


\subsection{Choosing the number of new edges}
\label{sec:choosing-num-of-new-edges} 

Certainly, the most important parameter of the proposed FACO algorithm is the
number of edges, $\textit{min\_new\_edges}$, that differentiate newly constructed
solutions from the \emph{source solution}. In the experiments, the probability
of using the best so far solution as the source solution was 1\%. Otherwise, it
was identical to the best solution from the previous iteration. In order to
check how the parameter influences the results, we chose four TSP instances
from the TSPLIB repository, namely \emph{d2103}, \emph{nrw1379},
\emph{pcb1173}, and \emph{rl1889}, and run the FACO with
$\textit{min\_new\_edges} \in [4, 6, \ldots, 20, 40, \ldots , 200 ]$. These values were selected
so that the differences between the base FACO and the FACO combined with the
2-opt LS are clearly visible. The other ACO-related parameters were as follows:
the number of ants was 128 in all cases, the number of iterations equaled 5000,
while the pheromone retention rate $\rho$ equaled $0.9$ for the base FACO, and
0.5 for the FACO with the LS.

\begin{figure}[]
    \begin{center}
        \input{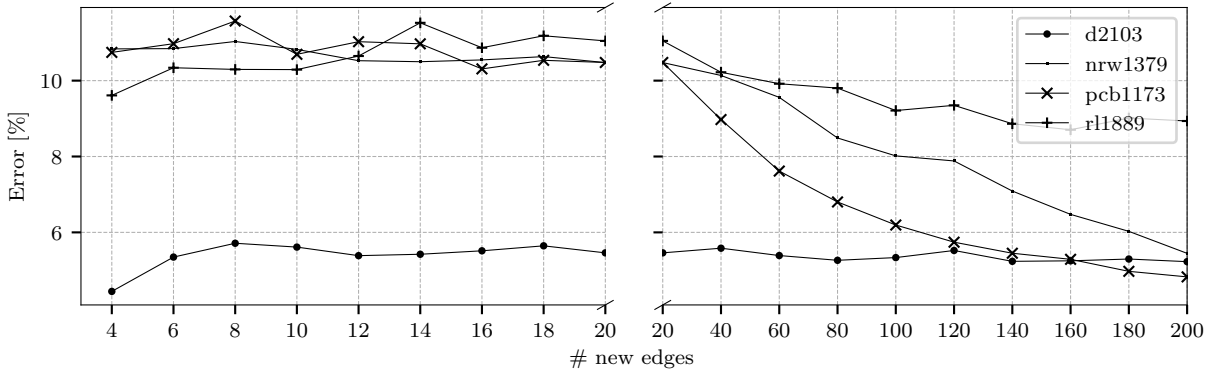}
    \end{center}
    \caption{Mean relative error of the FACO vs. the minimum number of new
    edges differentiating the newly constructed solutions from the
    current $\emph{source solution}$.}
    \label{fig:new-edges}
\end{figure}

Figure~\ref{fig:new-edges} shows the results for the FACO without the LS. As
can be seen, for three out of four considered instances, the results were
better if the algorithm was given greater freedom in constructing new
solutions. In contrast, allowing only a small number (no more than 20) of new
edges worsened the quality of the solution as only a tiny fraction of a
solution could change relative to the source solution. This observation agrees
with the intuition that the ACO (without a local search) is more
exploration-oriented and can make bigger jumps in the solution search
space~\cite{Stutzle2000, Dorigo2004}. Nevertheless, making small steps can be
beneficial in some cases, as can be observed for \emph{d2103} instance.
However, the difference in the relative error for the smallest and highest
values of $\textit{min\_new\_edges}$ was small, i.e., $4.4\%$ vs. $5.2\%$.

\begin{figure}[h]
    \begin{center}
        \input{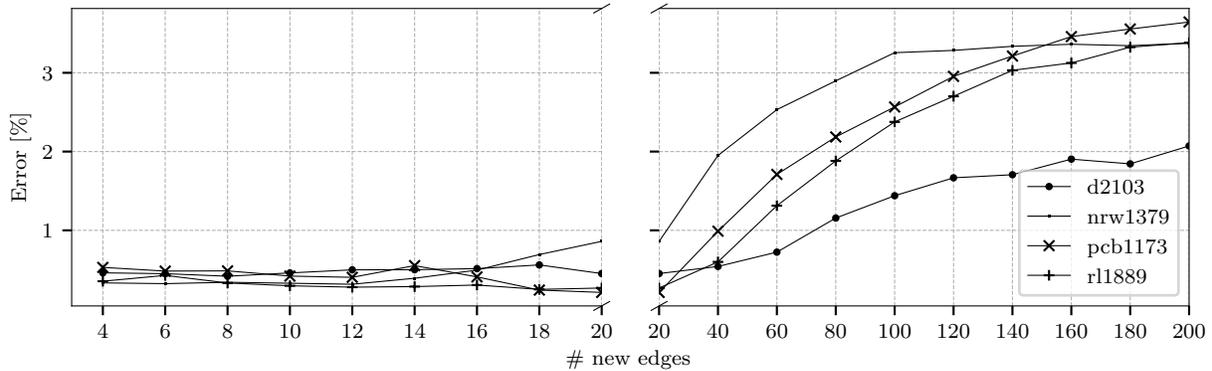}
    \end{center}
    \caption{Mean relative error of the FACO+LS vs. the minimum number of new
    edges differentiating the newly constructed solutions from the
    current $\emph{source solution}$.}
    \label{fig:new-edges-ls}
\end{figure}

The situation gets reversed if we pair the FACO with the 2-opt LS, as can be
seen in Fig.~\ref{fig:new-edges-ls}. Allowing only a small number of new
solution components (less than $20$) keeps the constructed solutions close to
the current \emph{source solution} while being sufficient for the LS to find
improving moves. As expected, the LS significantly improves the quality of the
results, which differ on average about 0.5\% from the optima.

\begin{figure}[h]
    \begin{center}
        \input{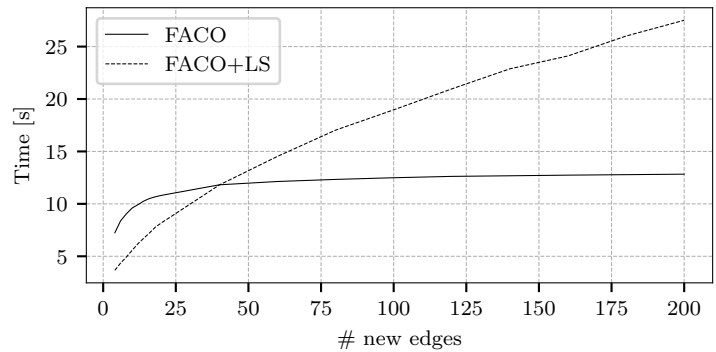}
    \end{center}
    \caption{Mean relative error of the FACO vs. the minimum number of new
    edges differentiating the newly constructed solutions from the
    current $\emph{source solution}$.}
    \label{fig:ls-nols-new-edges-time-cmp}
\end{figure}

Additionally, keeping the number of new edges in the constructed solutions
small is also beneficial for the computation time, especially if the LS is
applied.  Surprisingly, the FACO+LS can be faster than the base FACO as can be
seen in Fig.~\ref{fig:ls-nols-new-edges-time-cmp}. A possible explanation for
the phenomenon arises if we consider that the LS searches for improving moves
only among the (new) edges differentiating the constructed solution from the
source solution.  The higher the number of new edges in the constructed
solutions, the longer it takes to complete the FACO+LS, which becomes
significantly slower than the base FACO.

\subsection{Adjusting Pheromone Retention}
\label{sec:adjusting-pheromone-retention} 

Pheromone memory is an essential component of the ACO. It allows transferring
knowledge between successive iterations by increasing the pheromone
concentration on the edges of the current global (or iteration) best solution.
At the same time, the pheromone concentration lowers due to the evaporation
process as specified by Eq.~(\ref{eq:pher-update}). The amount of the pheromone
retained depends on the parameter $\rho$.

\begin{figure}
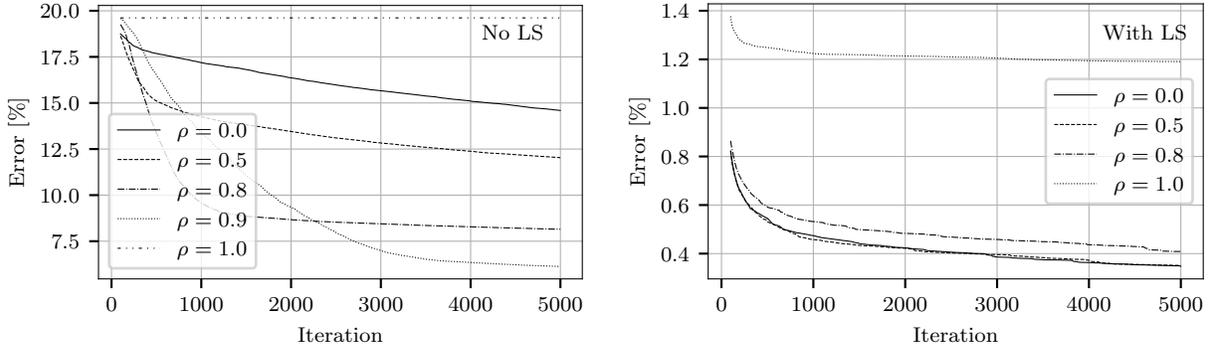

\centering
\begin{subfigure}{.5\textwidth}
  \centering
    \input{fig/rho-convergence-0.pgf}
\end{subfigure}%
\begin{subfigure}{.5\textwidth}
  \centering
    \input{fig/rho-convergence-1.pgf}
\end{subfigure}
    \caption{Convergence of the FACO in terms of the mean error relative to the
    optima. The lines correspond to the mean error values obtained for the TSP
    instances: \emph{d2103}, \emph{nrw1379}, \emph{pcb1173}, and
    \emph{rl1889}.}
\label{fig:rho-convergence}
\end{figure}

\begin{figure}
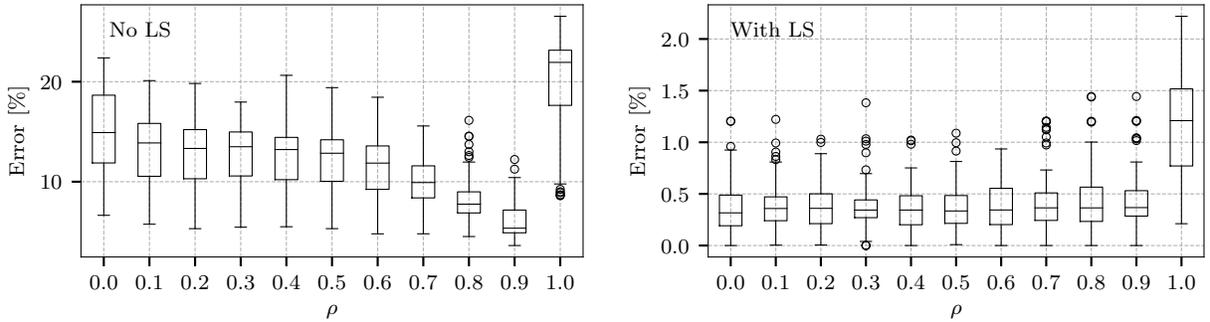

\centering
\begin{subfigure}{.5\textwidth}
  \centering
    \input{fig/paper-rho-cmp-0.pgf}
\end{subfigure}%
\begin{subfigure}{.5\textwidth}
  \centering
    \input{fig/paper-rho-cmp-1.pgf}
\end{subfigure}
\caption{
    Boxplots of the mean (final) solution error vs. the value of $\rho$
    parameter observed for the base FACO (left) and the FACO with the LS
    (right).
}
\label{fig:error-vs-rho}
\end{figure}

Figure~\ref{fig:rho-convergence} shows how the mean error of the current best
solution changes for a few selected $\rho$ values, while
Fig.~\ref{fig:error-vs-rho} shows the final error for the four TSP instances
considered previously.  The worst results were consistently obtained when $\rho
= 1$ both for the base FACO and the FACO with the 2-opt LS.  Setting $\rho$ to
1 prevents any pheromone from evaporating what means that all pheromone trails
have the same maximum value all the time. In such a scenario, the pheromone
does not affect the solution construction process, guided only by the heuristic
information. Additionally, in this case, if no LS is used, then the algorithm
easily gets trapped in a local minimum and has very little chance of escaping
it. The version with the LS shows a very slow convergence, but the final
results are much worse than for $\rho$ values smaller than $1$.

Another extreme can be observed if $\rho = 0$, in which case the previous
pheromone evaporates entirely. Only the edges belonging to the current global
(iteration) best solution have a non-minimum value.  This setting allows
obtaining better results proving that the pheromone memory is beneficial for
the algorithm's performance.  The best results for the FACO were obtained for
$\rho=0.9$ what agrees with the ACO behavior described in the
literature~\cite{Stutzle2000, Dorigo2004, Skinderowicz2020}.  Much more
interesting results were obtained for the FACO combined with the LS. For all
$\rho$ values lower than 1.0, the mean solution error was below 0.5\%, although
marginally better results were obtained for $\rho \le 0.5$.  This observation
suggests that the LS eliminates the need for a "long-term" pheromone memory
preferring more swift changes of the pheromone trails.


\subsection{Setting Number of Ants}
\label{sec:setting-num-of-ants} 

\begin{figure}[h]
    \begin{center}
        \input{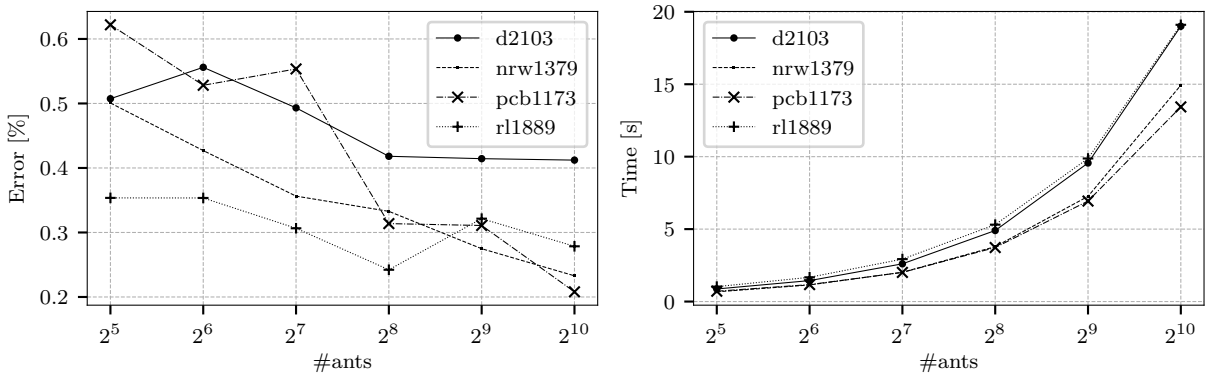}
    \end{center}
    \caption{Mean relative error (left) and mean computation time (right) of
    the FACO+LS vs. the number of ants.}
    \label{fig:var-ants-num}
\end{figure}

The number of ants in the FACO corresponds directly to the number of
constructed solutions as each ant constructs a single solution in every
iteration of the algorithm.
\R{
In order to avoid confusion, we have focused here only on the FACO+LS variant
following the observation that the ACO-based algorithms are typically used in
tandem with an efficient, problem-specific LS.  However, we can expect that the
probability of obtaining better outcomes increases as more solutions are being
constructed, regardless of the LS application. As the number of ants directly
affects the quality of the results and the execution time, the goal is to find
a setting that allows satisfactory outcomes to be found relatively quickly.
}
To check precisely how the number of ants affects the quality of the results, we have run the
FACO+LS for the same four TSP instances as
in~Sec.~\ref{sec:choosing-num-of-new-edges} and with the number of ants
increasing exponentially, i.e., $m \in \{ 32, 64, 128, 256, 512, 1024 \}$.  The
values of the other parameters were as follows: $\rho = 0.5$,
$\textit{min\_new\_edges} = 8$, and the number of iterations set to 3000.
Figure~\ref{fig:var-ants-num} shows how the mean solution error and the
computation time change with the number of ants.

As expected, the quality of the results improved with the increasing number of
constructed solutions (ants). The most notable improvement is visible when the
number of ants grows from 32 to 256, while the further increase to 1024 has a
much smaller impact. For example, quadrupling the number of ants from $2^8$ to
$2^{10}$ lowers the average error from 0.418\% to 0.412\% for \emph{d2103}
instance. On the other hand, the relation between the number of ants and the
computation time is roughly linear, i.e., doubling the number of ants doubles
the overall computation time. In all cases, the lengths of the resulting tours
were less than 1\% from the optima suggesting that the algorithm can be applied
even if the computation budget is modest.


\subsection{Performance evaluation}
\label{sec:performance-evaluation}

The strict control over how much a constructed solution can
differ from a selected previous (source) solution in the proposed FACO allows
shortening the computation time. On the other hand, it can increase the
difficulty of finding good-quality solutions.  For this reason, the first part
of the computational experiments is focused on comparison with other
high-performing ACOs on a set of TSP small to medium-sized instances from the
TSPLIB repository.  Based on the analysis presented
in~Sec.~\ref{sec:choosing-num-of-new-edges}--\ref{sec:setting-num-of-ants}, the
values of the FACO parameters were as follows: $\rho = 0.5$; the minimum number
of new edges, $\textit{min\_new\_edges}$, was set to 8; the number of
iterations was fixed to 5000. Finally, the number of ants, $m$, was set to $64
\lceil \frac{4 \sqrt{n}}{64} \rceil$, i.e. $4 \sqrt{n}$ rounded up to the
nearest multiple of 64, where $n$ denotes the size of the TSP instance.  This
setting was intended to offset the increase of the instance size as the largest
considered instance, \emph{d18512}, was 58 times larger than the smallest one,
\emph{lin318}.

\R{
Table~\ref{tab:cmp-esaco-mmas-rwm-bt}
compares the results of the proposed FACO
and two high-performing ACO-based algorithms, namely the ESACO
by~Ismkhan~\cite{Ismkhan2017} and the GPU-based
MMAS-RWM-BT~\cite{Skinderowicz2020} obtained for 15 TSP instances.  
The ESACO was implemented in C++ and executed on a computer with a 2 GHz Intel
CPU and 1 GB of RAM working under the Microsoft Windows 7 OS control. The
numbers of iterations and ants were set to 300 and 10, respectively.  The
MMAS-RWM-BT was implemented in C++, and the computations were carried out on a
machine with an HPC-grade Nvidia V100 GPU and a 2 GHz Intel Xeon CPU under the
control of Debian 9 OS. The main part of computations was executed in parallel
on the GPU.  The numbers of iterations and ants were 3000 and 800,
respectively.

As can be noticed, there is no clear winner between the three algorithms as
each algorithm excelled for instances with sizes in a particular range.  The
ESACO produced the best results for the few smallest instances, with sizes
between 318 and 1002 nodes. The ESACO is based on the ACS and employs an
effective LS comprising the 2-opt, 3-opt, and so-called double-bridge moves
what allows it to find high-quality solutions if the computation time is
sufficient. The MMAS and FACO rely on simpler but faster 2-opt LS, which may
not be enough to reach solutions very close to the optima.
The parallel MMAS was able to find the best quality solutions for four TSP
instances ranging from 1817 to 5915 nodes, with the mean error close to 0.5\%
relative to the optima. However, the error for the largest three instances was
much closer to 1\%, suggesting that more computation time was needed to keep
the quality closer to the previous levels.  Finally, the FACO produced the best
quality solutions for the five largest instances with 7397 to \numprint{18512}
nodes.  It is worth noting that the performance of the FACO was the most
consistent out of the three algorithms. In terms of the computation time, the
MMAS-RWM-BT and FACO were faster than the ESACO, and both came close for the
few smallest instances with up to 2392 nodes, while the FACO was significantly
faster for the larger instances. This result is noteworthy as the MMAS-RWM-BT
was executed on a powerful Nvidia V100 GPU while the FACO was executed on a
commodity CPU.  The good results obtained by the ESACO clearly show that the
effective LS can largely compensate for the slower (sequential) execution and
the smaller numbers of solutions generated.
}

\begin{table}[]
    \scriptsize
\centering
\caption{Comparison of the proposed FACO algorithm with the recently proposed
    high-performance ACO variants. Both mean and best tour lengths are reported.
    Errors relative to the best known results are given in brackets. The
    smallest mean tour lengths for each instance are marked in bold.}
\label{tab:cmp-esaco-mmas-rwm-bt}
\begin{tabular}{@{}lrrrrrrr@{}}
\toprule
\multicolumn{1}{c}{\multirow{2}{*}{\textbf{Instance}}} &
  \multicolumn{1}{c}{\multirow{2}{*}{\textbf{Optimum}}} &
    \multicolumn{2}{c}{\textbf{ESACO}~\cite{Ismkhan2017}} &
    \multicolumn{2}{c}{\textbf{MMAS-RWM-BT}~\cite{Skinderowicz2020}} &
    \multicolumn{2}{c}{\textbf{FACO}} \\ \cmidrule(lr){3-4} \cmidrule(lr){5-6} \cmidrule(l){7-8} 
\multicolumn{1}{c}{} &
  \multicolumn{1}{c}{} &
  \multicolumn{1}{c}{\textbf{Tour length}} &
  \multicolumn{1}{c}{\textbf{\begin{tabular}[c]{@{}c@{}}Time\\ {[}s{]}\end{tabular}}} &
  \multicolumn{1}{c}{\textbf{Tour length}} &
  \multicolumn{1}{c}{\textbf{\begin{tabular}[c]{@{}c@{}}Time\\ {[}s{]}\end{tabular}}} &
  \multicolumn{1}{c}{\textbf{Tour length}} &
  \multicolumn{1}{c}{\textbf{\begin{tabular}[c]{@{}c@{}}Time\\ {[}s{]}\end{tabular}}} \\ \hline
\textit{lin318} &
  42029 &
  \begin{tabular}[c]{@{}r@{}}\textbf{42053.9 (0.06\%)}\\ 42029.0 (0\%)\end{tabular} &
  10.1 &
  \begin{tabular}[c]{@{}r@{}}42069.6 (0.1\%)\\ 42029 (0\%)\end{tabular} &
  1.7 &
  \begin{tabular}[c]{@{}r@{}}42142 (0.27\%)\\ 42029 (0\%)\end{tabular} &
  1.9 \\ \midrule
\textit{pcb442} &
  50778 &
  \begin{tabular}[c]{@{}r@{}}\textbf{50803.6 (0.05\%)}\\ 50778.0 (0\%)\end{tabular} &
  11.5 &
  \begin{tabular}[c]{@{}r@{}}50950.7 (0.34\%)\\ 50809 (0.06\%)\end{tabular} &
  2.0 &
  \begin{tabular}[c]{@{}r@{}}50919 (0.28\%)\\ 50778 (0\%)\end{tabular} &
  2.0 \\ \midrule
\textit{att532} &
  27686 &
  \begin{tabular}[c]{@{}r@{}}\textbf{27701.2 (0\%)}\\ 27686 (0\%)\end{tabular} &
  23.1 &
  \begin{tabular}[c]{@{}r@{}}27708.9 (0.08\%)\\ 27686 (0\%)\end{tabular} &
  1.8 &
  \begin{tabular}[c]{@{}r@{}}27719 (0.12\%)\\ 27693 (0.03\%)\end{tabular} &
  2.2 \\ \midrule
\textit{rat783} &
  8806 &
  \begin{tabular}[c]{@{}r@{}}\textbf{8809.8 (0.04\%)}\\ 8806.0 (0\%)\end{tabular} &
  22.6 &
  \begin{tabular}[c]{@{}r@{}}8825.5 (0.22\%)\\ 8810 (0.05\%)\end{tabular} &
  3.1 &
  \begin{tabular}[c]{@{}r@{}}8831 (0.29\%)\\ 8810 (0.05\%)\end{tabular} &
  2.8 \\ \midrule
\textit{pr1002} &
  259045 &
  \begin{tabular}[c]{@{}r@{}}\textbf{259509.0 (0.18\%)}\\ 259045.0 (0\%)\end{tabular} &
  35.8 &
  \begin{tabular}[c]{@{}r@{}}259712.7 (0.26\%)\\ 259415 (0.14\%)\end{tabular} &
  4.0 &
  \begin{tabular}[c]{@{}r@{}}259858 (0.31\%)\\ 259078 (0.01\%)\end{tabular} &
  2.8 \\ \midrule
\textit{u1817} &
  57201 &
  - &
   &
  \begin{tabular}[c]{@{}r@{}}\textbf{57428.2 (0.4\%)}\\ 57209 (0.01\%)\end{tabular} &
  7.0 &
  \begin{tabular}[c]{@{}r@{}}57481 (0.49\%)\\ 57246 (0.08\%)\end{tabular} &
  5.6 \\ \midrule
\textit{pr2392} &
  378032 &
  - &
   &
  \begin{tabular}[c]{@{}r@{}}\textbf{379872.0 (0.49\%)}\\ 378390 (0.1\%)\end{tabular} &
  10.3 &
  \begin{tabular}[c]{@{}r@{}}379962 (0.51\%)\\ 378784 (0.20\%)\end{tabular} &
  9.5 \\ \midrule
\textit{fl3795} &
  28772 &
  \begin{tabular}[c]{@{}r@{}}28883.5 (0.39\%)\\ 28787.0 (0.05\%)\end{tabular} &
  119.3 &
  \begin{tabular}[c]{@{}r@{}}\textbf{28819.3 (0.16\%)}\\ 28793 (0.07\%)\end{tabular} &
  19.5 &
  \begin{tabular}[c]{@{}r@{}}28848 (0.26\%)\\ 28773 (0\%)\end{tabular} &
  12.2 \\ \midrule
\textit{fnl4461} &
  182566 &
  \begin{tabular}[c]{@{}r@{}}183446.0 (0.48\%)\\ 183254.0 (0.38\%)\end{tabular} &
  192.6 &
  \begin{tabular}[c]{@{}r@{}}183627.6 (0.58\%)\\ 183361 (0.44\%)\end{tabular} &
  34.9 &
  \begin{tabular}[c]{@{}r@{}}\textbf{183345 (0.43\%)}\\ 183153 (0.32\%)\end{tabular} &
  18.4 \\ \midrule
\textit{rl5915} &
  565530 &
  \begin{tabular}[c]{@{}r@{}}568935.0 (0.60\%)\\ 567177.0 (0.29\%)\end{tabular} &
  216.9 &
  \begin{tabular}[c]{@{}r@{}}\textbf{567699.9 (0.38\%)}\\ 566123 (0.11\%)\end{tabular} &
  49.3 &
  \begin{tabular}[c]{@{}r@{}}568104 (0.46\%)\\ 566566 (0.18\%)\end{tabular} &
  22.3 \\ \midrule
\textit{pla7397} &
  23260728 &
  \begin{tabular}[c]{@{}r@{}}23389341.0 (0.55\%)\\ 23345479.0 (0.36\%)\end{tabular} &
  213.9 &
  \begin{tabular}[c]{@{}r@{}}23386240.5 (0.54\%)\\ 23365046 (0.45\%)\end{tabular} &
  58.0 &
  \begin{tabular}[c]{@{}r@{}}\textbf{23377679 (0.50\%)}\\ 23326730 (0.28\%)\end{tabular} &
  29.2 \\ \midrule
\textit{rl11849} &
  923288 &
  \begin{tabular}[c]{@{}r@{}}930338.0 (0.76\%)\\ 928876.0 (0.61\%)\end{tabular} &
  575.8 &
  \begin{tabular}[c]{@{}r@{}}928192.1 (0.53\%)\\ 927452 (0.45\%)\end{tabular} &
  264.6 &
  \begin{tabular}[c]{@{}r@{}}\textbf{928166 (0.53\%)}\\ 926353 (0.33\%)\end{tabular} &
  54.8 \\ \midrule
\textit{usa13509} &
  19982859 &
  \begin{tabular}[c]{@{}r@{}}20195089.0 (1.06\%)\\ 20172735.0 (0.95\%)\end{tabular} &
  914.2 &
  \begin{tabular}[c]{@{}r@{}}20155797.4 (0.87\%)\\ 20127380 (0.72\%)\end{tabular} &
  269.1 &
  \begin{tabular}[c]{@{}r@{}}\textbf{20088662 (0.53\%)}\\ 20060245 (0.39\%)\end{tabular} &
  69.6 \\ \midrule
\textit{d15112} &
  1573084 &
  \begin{tabular}[c]{@{}r@{}}1589288.0 (1.03\%)\\ 1587150 (0.89\%)\end{tabular} &
  776.7 &
  \begin{tabular}[c]{@{}r@{}}1586604.05 (0.86\%)\\ 1584054 (0.70\%)\end{tabular} &
  404.2 &
  \begin{tabular}[c]{@{}r@{}}\textbf{1581127.6 (0.51\%)}\\ 1580192 (0.45\%)\end{tabular} &
  75.8 \\ \midrule
\textit{d18512} &
  645238 &
  \begin{tabular}[c]{@{}r@{}}653154.0 (1.23\%)\\ 652516.0 (1.13\%)\end{tabular} &
  684.4 &
  \begin{tabular}[c]{@{}r@{}}651730.5 (1.01\%)\\ 650961 (0.89\%)\end{tabular} &
  401.3 &
  \begin{tabular}[c]{@{}r@{}}\textbf{648833 (0.56\%)}\\ 648439 (0.50\%)\end{tabular} &
  100.1 \\ \bottomrule
\end{tabular}
\end{table}

\subsubsection{Large TSP instances}
\label{sec:large-tsp-instances}

\begin{figure}
\centering
\begin{subfigure}{.3\textwidth}
  \centering
  \includegraphics[width=0.9\linewidth]{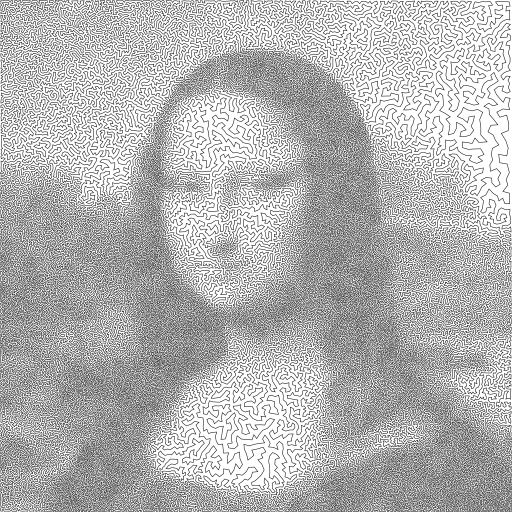}
    \caption{\emph{mona-lisa100K}}
\end{subfigure}%
\begin{subfigure}{.3\textwidth}
  \centering
  \includegraphics[width=0.9\linewidth]{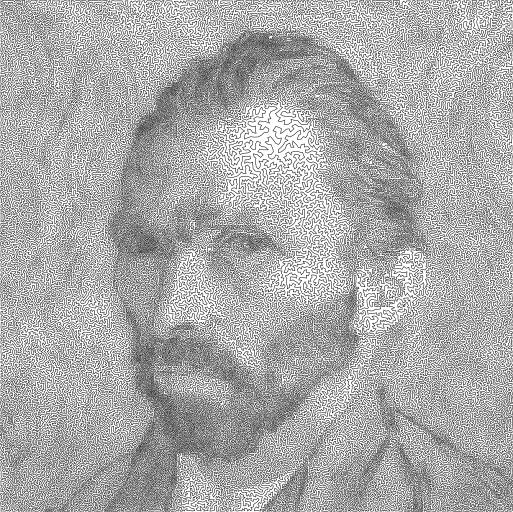}
  \caption{\emph{vangogh120K}}
\end{subfigure}%
\begin{subfigure}{.3\textwidth}
  \centering
  \includegraphics[width=0.9\linewidth]{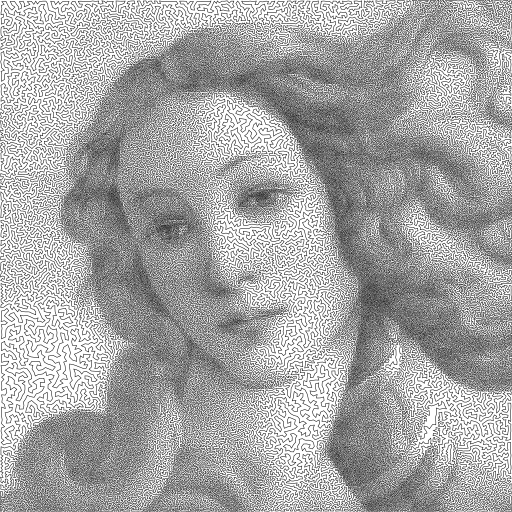}
  \caption{\emph{venus140K}}
\end{subfigure}
\begin{subfigure}{.3\textwidth}
  \centering
  \includegraphics[width=0.9\linewidth]{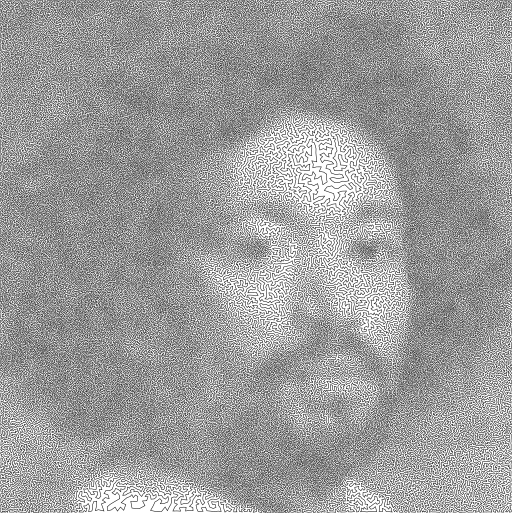}
  \caption{\emph{pareja160K}}
\end{subfigure}%
\begin{subfigure}{.3\textwidth}
  \centering
  \includegraphics[width=0.9\linewidth]{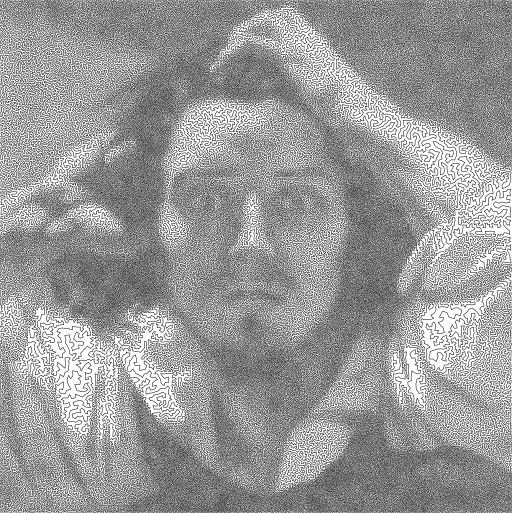}
  \caption{\emph{courbet180K}}
\end{subfigure}%
\begin{subfigure}{.3\textwidth}
  \centering
  \includegraphics[width=0.9\linewidth]{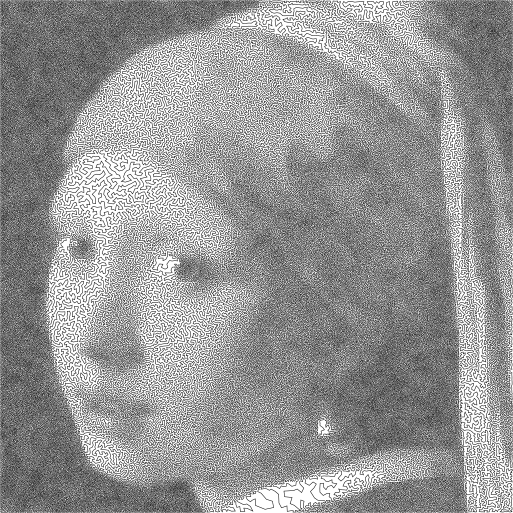}
    \caption{\emph{earring200K}}
\end{subfigure}
\caption{Visualisation of the best solutions obtained by the FACO for the
    TSP instances from \emph{TSP Art Instances}~\cite{ArtTSP}.}
\label{fig:best-art-tsp-solutions}
\end{figure}

The main motivations behind the changes introduced to the ACO in the proposed
FACO algorithm are in line with the recent research focus on enabling ACOs to
tackle large TSP instances with hundreds of thousands of
cities~\cite{Skinderowicz2013, Chitty2016, Peake2019, Martinez2021}.
Table~\ref{tab:art-tsp-cmp-with-acos} shows results of the ACO-RPMM algorithm
by~Peake et al.~\cite{Peake2019}, the PartialACO by~Chitty~\cite{Chitty2016},
and the proposed FACO for the instances from the \emph{TSP Art Instances}
dataset~\cite{ArtTSP}. 
The instances were created using techniques proposed
by~Bosch and Herman~\cite{Bosch2004} so that an optimum tour when drawn
resembles a selected classical painting, e.g., the \emph{mona-lisa100K}
instance corresponds to da Vinci's \emph{Mona Lisa} and has $10^5$ nodes
(cities). Figure~\ref{fig:best-art-tsp-solutions} depicts the best solutions
obtained by the FACO.

\begin{table}[]
    \small
\centering
\caption{Performance comparison of the ACO-based algorithms that were developed
    to tackle large TSP instances. The results are for the \emph{Art TSP}
    instances which have sizes ranging from 100k to 200k nodes~\cite{ArtTSP}.
    The FACO was run with the number of ants $m = 512$ and the number of
    iterations set to $10000$. The computations were repeated 30 times for each
    instance.}
\label{tab:art-tsp-cmp-with-acos}
\begin{tabular}{@{}lcccccrrr@{}}
\toprule
\multicolumn{1}{c}{\multirow{2}{*}{\textbf{Instance}}} &
  \multicolumn{1}{c}{\multirow{2}{*}{\textbf{\begin{tabular}[c]{@{}c@{}}Best
  known\\ tour~\cite{ArtTSP} \end{tabular}}}} &
      \multicolumn{2}{c}{\textbf{ACO-RPMM}~\cite{Peake2019} } &
    \multicolumn{2}{c}{\textbf{Partial ACO}~\cite{Chitty2016} } &
  \multicolumn{3}{c}{\textbf{FACO}} \\ \cmidrule(lr){3-4} \cmidrule(lr){5-6} \cmidrule(lr){7-9} 
\multicolumn{1}{c}{} &
  \multicolumn{1}{c}{} &
  \multicolumn{1}{c}{\begin{tabular}[c]{@{}c@{}}Error\\  {[}\%{]}\end{tabular}} &
  \multicolumn{1}{c}{\begin{tabular}[c]{@{}c@{}}Time\\ {[}h{]}\end{tabular}} &
  \multicolumn{1}{c}{\begin{tabular}[c]{@{}c@{}}Error\\  {[}\%{]}\end{tabular}} &
  \multicolumn{1}{c}{\begin{tabular}[c]{@{}c@{}}Time\\  {[}h{]}\end{tabular}} &
  \multicolumn{1}{c}{\begin{tabular}[c]{@{}c@{}}Avg. tour\\ length\end{tabular}} &
  \multicolumn{1}{c}{\begin{tabular}[c]{@{}c@{}}Error\\  {[}\%{]}\end{tabular}} &
  \multicolumn{1}{c}{\begin{tabular}[c]{@{}c@{}}Time\\  {[}h{]}\end{tabular}} \\ \midrule
\textit{mona-lisa100K} & 5757191 & 1.7 & 1.4 & 5.5 & 1.1 & 5793384 & 0.63 & 0.30  \\
\textit{vangogh120K}   & 6543609 & 1.8 & 1.9 & 5.8 & 1.5 & 6590084 & 0.71 & 0.39 \\
\textit{venus140K}     & 6810665 & 1.8 & 2.6 & 5.8 & 2.1 & 6861650 & 0.75 & 0.47 \\
\textit{pareja160K}    & 7619953 & 1.9 & 3.5 & -   & -   & 7681535 & 0.81 & 0.57 \\
\textit{courbet180K}   & 7888731 & 1.9 & 4.5 & -   & -   & 7958247 & 0.88 & 0.68 \\
\textit{earring200K}   & 8171677 & 2.0 & 6.0 & 7.2 & 5.1 & 8250696 & 0.96 & 0.86 \\ \bottomrule
\end{tabular}
\end{table}
\R{
All three tested algorithms had parallel implementations. The ACO-RPMM was
implemented using C++ and was tested on a computer with Intel Xeon E5-2640 CPU
with 20 cores (40 hardware threads) clocked at 2.4 GHz.  A single algorithm run
comprised 1000 iterations each, with 40 ants building their solutions.  The
PartialACO was tested on a computer with an Intel i7 CPU (4 cores), and the
SIMD instructions allowed to further speed up the node selection process. The
overall speedup reported by the authors was 30x to 40x over a plain, sequential
execution on the same machine. The number of ants was set to 16 while the
number of iterations was 10\,000.

As can be seen, the FACO outperformed the other two ACO-based algorithms in
terms of the quality of the final solutions and the execution time. The average
error of the best tour for each instance was below 1\% relative to the
best-published results~\cite{ArtTSP}, while the error of the ACO-RPMM ranged
from 1.7\% to 2.0\%, and the Partial ACO produced results differing by at least
5.5\%.  The difference to the ACO-RPMM is crucial as it produced the
second-best results in terms of quality. Moreover, the ACO-RPMM is also based
on the MMAS, and its implementation was parallel and executed on a slightly
more powerful computer.  Not surprisingly, in all cases, the quality of the
solutions drops with the growing size of the TSP instance tackled.
}

It is worth noting that the proposed FACO combines a general-purpose
metaheuristic (ACO) with a basic problem-specific LS (2-opt), which suggests
that a more sophisticated, TSP-oriented approach should outperform it easily.
Table~\ref{tab:lkh-cmp} compares results of the FACO with the LKH solver
by~Helsgaun~\cite{Helsgaun2009} which is the current state-of-the-art heuristic
for solving the TSP~\cite{Taillard2019, Taillard2021}. 
\R{The results were
obtained on the same machine (see~\ref{sec:Computing_environment} for the
details).} As expected, the LKH
can find solutions of very high quality, i.e., differing at most 0.035\% from
the best-known results~\cite{ArtTSP}, within a few hours of computation time.

\begin{table}[]
    \small
\centering
\caption{Comparison with the state-of-the-art LKH TSP solver.
    The parameters of the LKH were set following~\cite{Helsgaun2018Popmusic}.}
\label{tab:lkh-cmp}
\begin{tabular}{@{}lrrrrrr@{}}
\toprule
\multicolumn{1}{c}{\multirow{2}{*}{\textbf{Instance}}} & \multicolumn{3}{c}{\textbf{LKH}} & \multicolumn{3}{c}{\textbf{FACO}} \\
    \cmidrule(lr){2-4} \cmidrule(l){5-7} 
\multicolumn{1}{c}{} &
  \textbf{Tour length} &
  \textbf{Error {[}\%{]}} &
  \multicolumn{1}{c}{\textbf{Time {[}h{]}}} &
  \multicolumn{1}{c}{\textbf{\begin{tabular}[c]{@{}c@{}}Avg. tour\\  length\end{tabular}}} &
  \multicolumn{1}{l}{\textbf{Error {[}\%{]}}} &
  \multicolumn{1}{c}{\textbf{Time {[}h{]}}} \\ \midrule
\textit{mona-lisa100K}                                 & 5758679  & \textbf{0.028} & 3.62 & 5793384     & 0.631     & 0.30    \\
\textit{vangogh120K}                                   & 6545491  & \textbf{0.029} & 5.10 & 6590084     & 0.710     & 0.39    \\
\textit{venus140K}                                     & 6812492  & \textbf{0.026} & 4.00 & 6861650     & 0.749     & 0.47    \\
\textit{pareja160K}                                    & 7622309  & \textbf{0.031} & 5.20 & 7681535     & 0.808     & 0.57    \\
\textit{courbet180K}                                   & 7891308  & \textbf{0.033} & 7.51 & 7958247     & 0.881     & 0.68    \\
\textit{earring200K}                                   & 8174554  & \textbf{0.035} & 7.94 & 8250696     & 0.967     & 0.86    \\ \bottomrule
\end{tabular}
\end{table}

Interestingly, a more general approach can be advantageous in some cases,
making fewer assumptions about the problem.  Recently, Hougardy and
Zhong~\cite{Hougardy2021} have proposed a set of TSP instances explicitly
designed to be \emph{hard to solve}, especially for the exact TSP solvers like
\emph{Concorde} which is the fastest existing exact TSP solver~\cite{Concorde}.
Sizes of the instances range from several dozens to \numprint{100000} nodes which are all
points on a \emph{tetrahedron}. For example, when solving an instance with 200
nodes, it was noted that \emph{Concorde} was more than \numprint{1000000} times slower
than for similar-sized TSP instances from the TSPLIB repository. 
Table~\ref{tab:tnm-instances} shows results obtained by the FACO for the
hard-to-solve TSP instances with at least 500 nodes. The best-known results
obtained with the LKH solver are also reported. As can be noticed, the FACO was
able to find solutions of the same or even marginally better quality for a few
of the largest instances, i.e., with at least \numprint{20000} nodes.

\begin{table}[]
    \small
\centering
    \caption{Results for the \emph{hard to solve} TSP instances created by
    Hougardy and Zhong~\cite{Hougardy2021}. The best known results were
    obtained with the LKH solver~\cite{Helsgaun2018}. The improved results are
    shown in bold.} 
\label{tab:tnm-instances}
\R{
\begin{tabular}{@{}lrrrrr@{}}
\toprule

\multicolumn{1}{c}{\multirow{2}{*}{\textbf{Instance}}} &
    \multicolumn{1}{c}{\multirow{2}{*}{\textbf{Best known (LKH~\cite{Helsgaun2018})}}} &
  \multicolumn{4}{c}{\textbf{FACO}} \\ \cmidrule(l){3-6} 
\multicolumn{1}{c}{} &
  \multicolumn{1}{c}{} &
  \multicolumn{1}{c}{\textbf{Best}} &
  \multicolumn{1}{c}{\textbf{Worst}} &
  \multicolumn{1}{c}{\textbf{Mean}} &
  \multicolumn{1}{c}{\textbf{Time {[}s{]}}} \\ \midrule

\textit{Tnm502}    & 8\,749\,995      & 8\,749\,995               & 8\,749\,997      & 8\,749\,995.3      & 2.0    \\
\textit{Tnm601}    & 10\,615\,504     & 10\,615\,504              & 10\,615\,504     & 10\,615\,504.0     & 2.2    \\
\textit{Tnm700}    & 12\,488\,518     & 12\,488\,518              & 12\,488\,521     & 12\,488\,519.3     & 2.4    \\
\textit{Tnm802}    & 14\,377\,678     & 14\,377\,678              & 14\,377\,678     & 14\,377\,678.0     & 2.5    \\
\textit{Tnm901}    & 16\,256\,023     & 16\,256\,023              & 16\,256\,024     & 16\,256\,023.2     & 2.7    \\
\textit{Tnm1000}   & 18\,137\,296     & 18\,137\,298              & 18\,137\,298     & 18\,137\,298.0     & 2.9    \\
\textit{Tnm2002}   & 37\,029\,600     & 37\,029\,600              & 37\,029\,601     & 37\,029\,600.2     & 6.3    \\
\textit{Tnm3001}   & 55\,939\,349     & 55\,939\,349              & 55\,939\,352     & 55\,939\,349.5     & 10.1   \\
\textit{Tnm4000}   & 74\,858\,233     & 74\,858\,233              & 74\,858\,233     & 74\,858\,233.0     & 12.2   \\
\textit{Tnm5002}   & 93\,784\,081     & 93\,784\,081              & 93\,784\,081     & 93\,784\,081.0     & 17.3   \\
\textit{Tnm6001}   & 112\,708\,118    & 112\,708\,118             & 112\,708\,118    & 112\,708\,118.0    & 19.9   \\
\textit{Tnm7000}   & 131\,633\,371    & 131\,633\,371             & 131\,633\,374    & 131\,633\,372.0    & 26.1   \\
\textit{Tnm8002}   & 150\,561\,446    & 150\,561\,446             & 150\,561\,452    & 150\,561\,446.4    & 29.0   \\
\textit{Tnm9001}   & 169\,487\,546    & 169\,487\,546             & 169\,487\,548    & 169\,487\,546.1    & 31.9   \\
\textit{Tnm10000}  & 188\,414\,262    & 188\,414\,262             & 188\,414\,284    & 188\,414\,268.3    & 40.3   \\
\textit{Tnm20002}  & 377\,692\,238    & \textbf{377\,692\,219}    & 377\,692\,257    & 377\,692\,235.5    & 100.6  \\
\textit{Tnm30001}  & 566\,973\,296    & \textbf{566\,973\,186}    & 566\,975\,406    & 566\,973\,503.0    & 187.3  \\
\textit{Tnm40000}  & 756\,254\,243    & \textbf{756\,254\,121}    & 756\,257\,737    & 756\,254\,389.0    & 291.3  \\
\textit{Tnm50002}  & 945\,539\,807    & \textbf{945\,535\,600}    & 945\,535\,699    & 945\,535\,635.3    & 402.9  \\
\textit{Tnm60001}  & 1\,134\,820\,740 & \textbf{1\,134\,816\,445} & 1\,134\,816\,532 & 1\,134\,816\,474.6 & 546.6  \\
\textit{Tnm70000}  & 1\,324\,101\,816 & \textbf{1\,324\,097\,819} & 1\,324\,097\,927 & 1\,324\,097\,866.1 & 691.8  \\
\textit{Tnm80002}  & 1\,513\,392\,208 & \textbf{1\,513\,381\,620} & 1\,513\,391\,253 & 1\,513\,383\,905.6 & 797.4  \\
\textit{Tnm90001}  & 1\,702\,667\,051 & \textbf{1\,702\,662\,758} & 1\,702\,662\,910 & 1\,702\,662\,811.0 & 993.5  \\
\textit{Tnm100000} & 1\,891\,945\,975 & \textbf{1\,891\,945\,653} & 1\,891\,945\,678 & 1\,891\,945\,662.5 & 1271.5 \\ \bottomrule

\end{tabular}
}
\end{table}

\section{Conclusions}
\label{sec:Conclusions}

In this paper, we have shown that solving large-scale TSP instances efficiently
with the ACO-based algorithm (MMAS) is possible with the help of a few
relatively simple changes to the base algorithm and careful incorporation of a
TSP-specific local search (2-opt).  The main change concerns the constructive
nature of the ACO, which involves building solutions to the problem from
scratch in every iteration of the algorithm~\cite{Dorigo1996}.  This approach
contrasts perturbation-based heuristics, which apply several (typically small)
modifications to the existing solutions, e.g., found in the previous
iterations~\cite{Helsgaun2000}. The latter approach is typically less
time-consuming, particularly as the size of the problem increases. The proposed
FACO follows a typical constructive behavior of the ACO but only until it
selects a predefined number of solution components (edges) that are not present
in the \emph{source solution}. Next, the remaining portion of the solution is
copied from the source solution. This strategy allows the FACO to gain the
speed benefits of the perturbation-based approach without losing the
constructive nature of the ACO.  Moreover, keeping track of the differences
between the constructed and the source solutions allows for a more efficient
integration with the problem-specific LS (2-opt), which can focus only on the recently
introduced components and skip the parts that did not
change.

The proposed FACO demonstrates a robust performance when solving TSP instances
with dozens to hundreds of thousands of nodes.  For example, the algorithm
required less than an hour to find solutions within 1\% from the best-known
results for the \emph{TSP Art Instances}~\cite{ArtTSP} with 100k to 200k nodes
while running on a computer with an 8-core CPU. This can be seen as an
improvement over the recently proposed ACO-based approaches, including the
ACO-RPMM~\cite{Peake2019}, the PartialACO~\cite{Chitty2016}, and the
ACOTSP-MF~\cite{Martinez2021}. As such, the FACO is a valuable addition to the
family of ACO-based algorithms.  Not surprisingly, the FACO cannot compete with
the state-of-the-art LKH solver when solving the TSP instances from the TSPLIB
or TSP Art datasets. However, due to making fewer assumptions about the
problem, it is able to find the same or even slightly better solutions for the
hard-to-solve TSP instances by~Hougardy and Zhong~\cite{Hougardy2021}.

\subsection*{Future work}

The presented work may be extended in multiple directions.  The proposed method
of controlling the extent to which a new solution constructed by an ant differs
from a selected previous (source) solution is largely problem-agnostic while
being easy to implement at the same time. For these reasons, the method could
prove valuable within contexts of different optimization problems. On the other
hand, it should be possible to improve further the efficiency of the proposed
FACO in the context of the TSP with the help of a more sophisticated LS, e.g.,
one considering also non-sequential moves~\cite{Helsgaun2000}. Another
direction points to a potential reduction of the FACO computation time with the
help of the GPUs, or vector instructions (SIMD) offered by modern CPUs.


\bibliographystyle{plainnat}
\footnotesize
\bibliography{article}

\begin{thebibliography}{48}
\providecommand{\natexlab}[1]{#1}
\providecommand{\url}[1]{\texttt{#1}}
\expandafter\ifx\csname urlstyle\endcsname\relax
  \providecommand{\doi}[1]{doi: #1}\else
  \providecommand{\doi}{doi: \begingroup \urlstyle{rm}\Url}\fi

\bibitem[Applegate(2011)]{Applegate2011}
David Applegate.
\newblock \emph{The Traveling Salesman Problem : a Computational Study}.
\newblock Princeton University Press, Princeton, 2011.
\newblock ISBN 978-0-691-12993-8.

\bibitem[Applegate et~al.()Applegate, Bixby, Chv{\'a}tal, and Cook]{Concorde}
David Applegate, Robert Bixby, Vasek Chv{\'a}tal, and William Cook.
\newblock Concorde.
\newblock URL
  \url{http://www.math.uwaterloo.ca/tsp/concorde/downloads/downloads.htm}.

\bibitem[Bell and McMullen(2004)]{Bell2004}
John~E. Bell and Patrick~R. McMullen.
\newblock Ant colony optimization techniques for the vehicle routing problem.
\newblock \emph{Adv. Eng. Informatics}, 18\penalty0 (1):\penalty0 41--48, 2004.
\newblock \doi{10.1016/j.aei.2004.07.001}.
\newblock URL \url{https://doi.org/10.1016/j.aei.2004.07.001}.

\bibitem[Bentley(1992)]{Bentley1992}
Jon~Louis Bentley.
\newblock Fast algorithms for geometric traveling salesman problems.
\newblock \emph{{INFORMS} Journal on Computing}, 4\penalty0 (4):\penalty0
  387--411, 1992.
\newblock \doi{10.1287/ijoc.4.4.387}.
\newblock URL \url{https://doi.org/10.1287/ijoc.4.4.387}.

\bibitem[Bosch and Herman(2004)]{Bosch2004}
Robert Bosch and Adrianne Herman.
\newblock Continuous line drawings via the traveling salesman problem.
\newblock \emph{Oper. Res. Lett.}, 32\penalty0 (4):\penalty0 302--303, 2004.
\newblock \doi{10.1016/j.orl.2003.10.001}.
\newblock URL \url{https://doi.org/10.1016/j.orl.2003.10.001}.

\bibitem[Cecilia et~al.(2013)Cecilia, Garc{\'{\i}}a, Nisbet, Amos, and
  Ujaldon]{Cecilia2012}
Jos{\'{e}}~M. Cecilia, Jos{\'{e}}~M. Garc{\'{\i}}a, Andy Nisbet, Martyn Amos,
  and Manuel Ujaldon.
\newblock Enhancing data parallelism for ant colony optimization on gpus.
\newblock \emph{J. Parallel Distrib. Comput.}, 73\penalty0 (1):\penalty0
  42--51, 2013.
\newblock \doi{10.1016/j.jpdc.2012.01.002}.
\newblock URL \url{https://doi.org/10.1016/j.jpdc.2012.01.002}.

\bibitem[Cecilia et~al.(2018)Cecilia, Llanes, Abell{\'{a}}n,
  G{\'{o}}mez{-}Luna, Chang, and Hwu]{Cecilia2018}
Jos{\'{e}}~M. Cecilia, Antonio Llanes, Jos{\'{e}}~L. Abell{\'{a}}n, Juan
  G{\'{o}}mez{-}Luna, Li{-}Wen Chang, and Wen{-}Mei~W. Hwu.
\newblock High-throughput ant colony optimization on graphics processing units.
\newblock \emph{J. Parallel Distrib. Comput.}, 113:\penalty0 261--274, 2018.
\newblock \doi{10.1016/j.jpdc.2017.12.002}.
\newblock URL \url{https://doi.org/10.1016/j.jpdc.2017.12.002}.

\bibitem[Chitty(2017)]{Chitty2016}
Darren~M. Chitty.
\newblock Applying {ACO} to large scale {TSP} instances.
\newblock \emph{CoRR}, abs/1709.03187, 2017.
\newblock URL \url{http://arxiv.org/abs/1709.03187}.

\bibitem[Choong et~al.(2019)Choong, Wong, and Lim]{Choong2019}
Shin~Siang Choong, Li{-}Pei Wong, and Chee~Peng Lim.
\newblock An artificial bee colony algorithm with a modified choice function
  for the traveling salesman problem.
\newblock \emph{Swarm and Evolutionary Computation}, 44:\penalty0 622--635,
  2019.
\newblock \doi{10.1016/j.swevo.2018.08.004}.
\newblock URL \url{https://doi.org/10.1016/j.swevo.2018.08.004}.

\bibitem[Colorni et~al.(1991)Colorni, Dorigo, Maniezzo, et~al.]{Colorni1991}
Alberto Colorni, Marco Dorigo, Vittorio Maniezzo, et~al.
\newblock Distributed optimization by ant colonies.
\newblock In \emph{Proceedings of the first European conference on artificial
  life}, volume 142, pages 134--142. Paris, France, 1991.

\bibitem[Cook()]{ArtTSP}
William Cook.
\newblock {TSP Art Instances}.
\newblock URL \url{https://www.math.uwaterloo.ca/tsp/data/art}.

\bibitem[Del{\'e}vacq et~al.(2013)Del{\'e}vacq, Delisle, Gravel, and
  Krajecki]{Delevacq2013}
Audrey Del{\'e}vacq, Pierre Delisle, Marc Gravel, and Micha{\"{e}}l Krajecki.
\newblock Parallel ant colony optimization on graphics processing units.
\newblock \emph{J. Parallel Distrib. Comput.}, 73\penalty0 (1):\penalty0
  52--61, 2013.
\newblock \doi{10.1016/j.jpdc.2012.01.003}.
\newblock URL \url{https://doi.org/10.1016/j.jpdc.2012.01.003}.

\bibitem[Deng et~al.(2019)Deng, Xu, and Zhao]{Deng2019}
Wu~Deng, Junjie Xu, and Huimin Zhao.
\newblock An improved ant colony optimization algorithm based on hybrid
  strategies for scheduling problem.
\newblock \emph{{IEEE} Access}, 7:\penalty0 20281--20292, 2019.
\newblock \doi{10.1109/ACCESS.2019.2897580}.
\newblock URL \url{https://doi.org/10.1109/ACCESS.2019.2897580}.

\bibitem[D{\"{o}}keroglu et~al.(2019)D{\"{o}}keroglu, Sevin{\c{c}},
  Kucukyilmaz, and Cosar]{Dokeroglu2019}
Tansel D{\"{o}}keroglu, Ender Sevin{\c{c}}, Tayfun Kucukyilmaz, and Ahmet
  Cosar.
\newblock A survey on new generation metaheuristic algorithms.
\newblock \emph{Comput. Ind. Eng.}, 137, 2019.
\newblock \doi{10.1016/j.cie.2019.106040}.
\newblock URL \url{https://doi.org/10.1016/j.cie.2019.106040}.

\bibitem[Dorigo and Gambardella(1997)]{Dorigo1997}
Marco Dorigo and Luca~Maria Gambardella.
\newblock Ant colony system: a cooperative learning approach to the traveling
  salesman problem.
\newblock \emph{{IEEE} Trans. Evol. Comput.}, 1\penalty0 (1):\penalty0 53--66,
  1997.
\newblock \doi{10.1109/4235.585892}.
\newblock URL \url{https://doi.org/10.1109/4235.585892}.

\bibitem[Dorigo and St{\"{u}}tzle(2004)]{Dorigo2004}
Marco Dorigo and Thomas St{\"{u}}tzle.
\newblock \emph{Ant colony optimization}.
\newblock {MIT} Press, 2004.
\newblock ISBN 978-0-262-04219-2.
\newblock \doi{10.7551/mitpress/1290.001.0001}.
\newblock URL \url{https://doi.org/10.7551/mitpress/1290.001.0001}.

\bibitem[Dorigo et~al.(1996)Dorigo, Maniezzo, and Colorni]{Dorigo1996}
Marco Dorigo, Vittorio Maniezzo, and Alberto Colorni.
\newblock Ant system: optimization by a colony of cooperating agents.
\newblock \emph{{IEEE} Trans. Systems, Man, and Cybernetics, Part {B}},
  26\penalty0 (1):\penalty0 29--41, 1996.
\newblock \doi{10.1109/3477.484436}.
\newblock URL \url{https://doi.org/10.1109/3477.484436}.

\bibitem[Engelbrecht(2005)]{Engelbrecht2005}
Andries~Petrus Engelbrecht.
\newblock \emph{Fundamentals of Computational Swarm Intelligence}.
\newblock Wiley, 2005.
\newblock ISBN 978-0-470-09191-3.
\newblock URL
  \url{http://eu.wiley.com/WileyCDA/WileyTitle/productCd-0470091916.html}.

\bibitem[Gambardella et~al.(2012)Gambardella, Montemanni, and
  Weyland]{Gambardella2012}
L.M. Gambardella, R.~Montemanni, and D.~Weyland.
\newblock Coupling ant colony systems with strong local searches.
\newblock \emph{European Journal of Operational Research}, 220\penalty0
  (3):\penalty0 831--843, 2012.
\newblock ISSN 0377-2217.
\newblock \doi{https://doi.org/10.1016/j.ejor.2012.02.038}.
\newblock URL
  \url{https://www.sciencedirect.com/science/article/pii/S0377221712001889}.

\bibitem[Guntsch and Middendorf(2002)]{Guntsch2002}
Michael Guntsch and Martin Middendorf.
\newblock A population based approach for {ACO}.
\newblock In Stefano Cagnoni, Jens Gottlieb, Emma Hart, Martin Middendorf, and
  G{\"{u}}nther~R. Raidl, editors, \emph{Applications of Evolutionary
  Computing, EvoWorkshops 2002: EvoCOP, EvoIASP, EvoSTIM/EvoPLAN, Kinsale,
  Ireland, April 3-4, 2002, Proceedings}, volume 2279 of \emph{Lecture Notes in
  Computer Science}, pages 72--81. Springer, 2002.
\newblock \doi{10.1007/3-540-46004-7\_8}.
\newblock URL \url{https://doi.org/10.1007/3-540-46004-7\_8}.

\bibitem[Helsgaun(2000)]{Helsgaun2000}
Keld Helsgaun.
\newblock An effective implementation of the lin-kernighan traveling salesman
  heuristic.
\newblock \emph{Eur. J. Oper. Res.}, 126\penalty0 (1):\penalty0 106--130, 2000.
\newblock \doi{10.1016/S0377-2217(99)00284-2}.
\newblock URL \url{https://doi.org/10.1016/S0377-2217(99)00284-2}.

\bibitem[Helsgaun(2009)]{Helsgaun2009}
Keld Helsgaun.
\newblock General \emph{k}-opt submoves for the lin-kernighan {TSP} heuristic.
\newblock \emph{Math. Program. Comput.}, 1\penalty0 (2-3):\penalty0 119--163,
  2009.
\newblock \doi{10.1007/s12532-009-0004-6}.
\newblock URL \url{https://doi.org/10.1007/s12532-009-0004-6}.

\bibitem[Helsgaun(2018{\natexlab{a}})]{Helsgaun2018}
Keld Helsgaun.
\newblock {Best LKH Solutions for Tnm Instances}, 2018{\natexlab{a}}.
\newblock URL \url{http://webhotel4.ruc.dk/~keld/research/LKH/Best LKH
  solutions for Tnm instances.pdf}.

\bibitem[Helsgaun(2018{\natexlab{b}})]{Helsgaun2018Popmusic}
Keld Helsgaun.
\newblock Using {POPMUSIC} for candidate set generation in the
  lin-kernighan-helsgaun {TSP} solver.
\newblock Technical report, Department of Computer Science, Roskilde
  University, {DK-4000} Roskilde, Denmark, July 2018{\natexlab{b}}.
\newblock URL
  \url{http://webhotel4.ruc.dk/~keld/research/LKH/POPMUSIC_REPORT.pdf}.

\bibitem[Hougardy and Zhong(2021)]{Hougardy2021}
Stefan Hougardy and Xianghui Zhong.
\newblock Hard to solve instances of the euclidean traveling salesman problem.
\newblock \emph{Math. Program. Comput.}, 13\penalty0 (1):\penalty0 51--74,
  2021.
\newblock \doi{10.1007/s12532-020-00184-5}.
\newblock URL \url{https://doi.org/10.1007/s12532-020-00184-5}.

\bibitem[Ismkhan(2017)]{Ismkhan2017}
Hassan Ismkhan.
\newblock Effective heuristics for ant colony optimization to handle
  large-scale problems.
\newblock \emph{Swarm and Evolutionary Computation}, 32:\penalty0 140--149,
  2017.
\newblock \doi{10.1016/j.swevo.2016.06.006}.
\newblock URL \url{https://doi.org/10.1016/j.swevo.2016.06.006}.

\bibitem[Kennedy(2006)]{Kennedy2006}
James Kennedy.
\newblock Swarm intelligence.
\newblock In Albert~Y. Zomaya, editor, \emph{Handbook of Nature-Inspired and
  Innovative Computing - Integrating Classical Models with Emerging
  Technologies}, pages 187--219. Springer, 2006.
\newblock \doi{10.1007/0-387-27705-6\_6}.
\newblock URL \url{https://doi.org/10.1007/0-387-27705-6\_6}.

\bibitem[Leguizamon and Michalewicz(1999)]{Leguizamon1999}
Guillermo Leguizamon and Zbigniew Michalewicz.
\newblock A new version of ant system for subset problems.
\newblock In \emph{Proceedings of the 1999 Congress on Evolutionary
  Computation-CEC99 (Cat. No. 99TH8406)}, volume~2, pages 1459--1464. IEEE,
  1999.

\bibitem[Lin(1965)]{Lin1965}
Shen Lin.
\newblock Computer solutions of the traveling salesman problem.
\newblock \emph{Bell System Technical Journal}, 44\penalty0 (10):\penalty0
  2245--2269, 1965.

\bibitem[Lin and Kernighan(1973)]{Lin1973}
Shen Lin and Brian~W. Kernighan.
\newblock An effective heuristic algorithm for the traveling-salesman problem.
\newblock \emph{Oper. Res.}, 21\penalty0 (2):\penalty0 498--516, 1973.
\newblock \doi{10.1287/opre.21.2.498}.
\newblock URL \url{https://doi.org/10.1287/opre.21.2.498}.

\bibitem[L{\'{o}}pez{-}Ib{\'{a}}{\~{n}}ez and Blum(2010)]{LopezIbanez2010}
Manuel L{\'{o}}pez{-}Ib{\'{a}}{\~{n}}ez and Christian Blum.
\newblock Beam-aco for the travelling salesman problem with time windows.
\newblock \emph{Comput. Oper. Res.}, 37\penalty0 (9):\penalty0 1570--1583,
  2010.
\newblock \doi{10.1016/j.cor.2009.11.015}.
\newblock URL \url{https://doi.org/10.1016/j.cor.2009.11.015}.

\bibitem[Marinakis et~al.(2011)Marinakis, Marinaki, and Dounias]{Marinakis2010}
Yannis Marinakis, Magdalene Marinaki, and Georgios Dounias.
\newblock Honey bees mating optimization algorithm for the euclidean traveling
  salesman problem.
\newblock \emph{Inf. Sci.}, 181\penalty0 (20):\penalty0 4684--4698, 2011.
\newblock \doi{10.1016/j.ins.2010.06.032}.
\newblock URL \url{https://doi.org/10.1016/j.ins.2010.06.032}.

\bibitem[Mart{\'{\i}}nez and Garc{\'{\i}}a(2021)]{Martinez2021}
Pablo~Antonio Mart{\'{\i}}nez and Jos{\'{e}}~M. Garc{\'{\i}}a.
\newblock {ACOTSP-MF:} {A} memory-friendly and highly scalable {ACOTSP}
  approach.
\newblock \emph{Eng. Appl. Artif. Intell.}, 99:\penalty0 104131, 2021.
\newblock \doi{10.1016/j.engappai.2020.104131}.
\newblock URL \url{https://doi.org/10.1016/j.engappai.2020.104131}.

\bibitem[Nezamabadi{-}pour et~al.(2006)Nezamabadi{-}pour, Saryazdi, and
  Rashedi]{Nezamabadi2006}
Hossein Nezamabadi{-}pour, Saeid Saryazdi, and Esmat Rashedi.
\newblock Edge detection using ant algorithms.
\newblock \emph{Soft Comput.}, 10\penalty0 (7):\penalty0 623--628, 2006.
\newblock \doi{10.1007/s00500-005-0511-y}.
\newblock URL \url{https://doi.org/10.1007/s00500-005-0511-y}.

\bibitem[Peake et~al.(2019)Peake, Amos, Yiapanis, and Lloyd]{Peake2019}
Joshua Peake, Martyn Amos, Paraskevas Yiapanis, and Huw Lloyd.
\newblock Scaling techniques for parallel ant colony optimization on large
  problem instances.
\newblock In Anne Auger and Thomas St{\"{u}}tzle, editors, \emph{Proceedings of
  the Genetic and Evolutionary Computation Conference, {GECCO} 2019, Prague,
  Czech Republic, July 13-17, 2019}, pages 47--54. {ACM}, 2019.
\newblock \doi{10.1145/3321707.3321832}.
\newblock URL \url{https://doi.org/10.1145/3321707.3321832}.

\bibitem[Pedemonte et~al.(2011)Pedemonte, Nesmachnow, and
  Cancela]{Pedemonte2011}
Mart{\'{\i}}n Pedemonte, Sergio Nesmachnow, and H{\'{e}}ctor Cancela.
\newblock A survey on parallel ant colony optimization.
\newblock \emph{Appl. Soft Comput.}, 11\penalty0 (8):\penalty0 5181--5197,
  2011.
\newblock \doi{10.1016/j.asoc.2011.05.042}.
\newblock URL \url{https://doi.org/10.1016/j.asoc.2011.05.042}.

\bibitem[Shmygelska and Hoos(2005)]{Shmygelska2005}
Alena Shmygelska and Holger~H. Hoos.
\newblock An ant colony optimisation algorithm for the 2d and 3d hydrophobic
  polar protein folding problem.
\newblock \emph{{BMC} Bioinform.}, 6:\penalty0 30, 2005.
\newblock \doi{10.1186/1471-2105-6-30}.
\newblock URL \url{https://doi.org/10.1186/1471-2105-6-30}.

\bibitem[Skinderowicz(2013)]{Skinderowicz2013}
Rafa{\l} Skinderowicz.
\newblock Ant colony system with selective pheromone memory for {SOP}.
\newblock In Costin Badica, Ngoc~Thanh Nguyen, and Marius Brezovan, editors,
  \emph{Computational Collective Intelligence. Technologies and Applications -
  5th International Conference, {ICCCI} 2013, Craiova, Romania, September
  11-13, 2013, Proceedings}, volume 8083 of \emph{Lecture Notes in Computer
  Science}, pages 711--720. Springer, 2013.
\newblock \doi{10.1007/978-3-642-40495-5\_71}.
\newblock URL \url{https://doi.org/10.1007/978-3-642-40495-5\_71}.

\bibitem[Skinderowicz(2016)]{Skinderowicz2016}
Rafa{\l} Skinderowicz.
\newblock The {GPU}-based parallel {A}nt {C}olony {S}ystem.
\newblock \emph{J. Parallel Distrib. Comput.}, 98:\penalty0 48--60, 2016.
\newblock \doi{10.1016/j.jpdc.2016.04.014}.
\newblock URL \url{https://doi.org/10.1016/j.jpdc.2016.04.014}.

\bibitem[Skinderowicz(2020)]{Skinderowicz2020}
Rafa{\l} Skinderowicz.
\newblock Implementing a {GPU}-based parallel {MAX-MIN} ant system.
\newblock \emph{Future Gener. Comput. Syst.}, 106:\penalty0 277--295, 2020.
\newblock \doi{10.1016/j.future.2020.01.011}.
\newblock URL \url{https://doi.org/10.1016/j.future.2020.01.011}.

\bibitem[St{\"{u}}tzle and Hoos(2000)]{Stutzle2000}
Thomas St{\"{u}}tzle and Holger~H. Hoos.
\newblock {MAX-MIN} ant system.
\newblock \emph{Future Generation Comp. Syst.}, 16\penalty0 (8):\penalty0
  889--914, 2000.
\newblock \doi{10.1016/S0167-739X(00)00043-1}.
\newblock URL \url{https://doi.org/10.1016/S0167-739X(00)00043-1}.

\bibitem[Taillard(2021)]{Taillard2021}
{\'E}ric~D Taillard.
\newblock A linearithmic heuristic for the travelling salesman problem.
\newblock \emph{European Journal of Operational Research}, 2021.

\bibitem[Taillard and Helsgaun(2019)]{Taillard2018}
{\'{E}}ric~D. Taillard and Keld Helsgaun.
\newblock {POPMUSIC} for the travelling salesman problem.
\newblock \emph{Eur. J. Oper. Res.}, 272\penalty0 (2):\penalty0 420--429, 2019.
\newblock \doi{10.1016/j.ejor.2018.06.039}.
\newblock URL \url{https://doi.org/10.1016/j.ejor.2018.06.039}.

\bibitem[Tirado et~al.(2017)Tirado, Barrientos, Gonz{\'{a}}lez, and
  Mora]{Tirado2017}
Felipe Tirado, Ricardo~J. Barrientos, Paulo Gonz{\'{a}}lez, and Marco Mora.
\newblock Efficient exploitation of the xeon phi architecture for the ant
  colony optimization {(ACO)} metaheuristic.
\newblock \emph{J. Supercomput.}, 73\penalty0 (11):\penalty0 5053--5070, 2017.
\newblock \doi{10.1007/s11227-017-2124-5}.
\newblock URL \url{https://doi.org/10.1007/s11227-017-2124-5}.

\bibitem[Yang et~al.(2016)Yang, Deb, Fong, He, and Zhao]{Yang2016}
Xin{-}She Yang, Suash Deb, Simon Fong, Xingshi He, and Yuxin Zhao.
\newblock From swarm intelligence to metaheuristics: Nature-inspired
  optimization algorithms.
\newblock \emph{Computer}, 49\penalty0 (9):\penalty0 52--59, 2016.
\newblock \doi{10.1109/MC.2016.292}.
\newblock URL \url{https://doi.org/10.1109/MC.2016.292}.

\bibitem[Zhong et~al.(2019)Zhong, Wang, Lin, and Zhang]{Zhong2018}
Yiwen Zhong, Lijin Wang, Min Lin, and Hui Zhang.
\newblock Discrete pigeon-inspired optimization algorithm with metropolis
  acceptance criterion for large-scale traveling salesman problem.
\newblock \emph{Swarm Evol. Comput.}, 48:\penalty0 134--144, 2019.
\newblock \doi{10.1016/j.swevo.2019.04.002}.
\newblock URL \url{https://doi.org/10.1016/j.swevo.2019.04.002}.

\bibitem[Zhou et~al.(2018)Zhou, He, Hou, and Qiu]{Zhou2018}
Yi~Zhou, Fazhi He, Neng Hou, and Yimin Qiu.
\newblock Parallel ant colony optimization on multi-core {SIMD} cpus.
\newblock \emph{Future Generation Comp. Syst.}, 79:\penalty0 473--487, 2018.
\newblock \doi{10.1016/j.future.2017.09.073}.
\newblock URL \url{https://doi.org/10.1016/j.future.2017.09.073}.

\bibitem[Éric D.~Taillard and Helsgaun(2019)]{Taillard2019}
Éric D.~Taillard and Keld Helsgaun.
\newblock {POPMUSIC} for the travelling salesman problem.
\newblock \emph{European Journal of Operational Research}, 272\penalty0
  (2):\penalty0 420--429, 2019.
\newblock ISSN 0377-2217.
\newblock \doi{https://doi.org/10.1016/j.ejor.2018.06.039}.
\newblock URL
  \url{https://www.sciencedirect.com/science/article/pii/S0377221718305745}.

\end{thebibliography}

\end{document}